\documentclass[Afour,sageh,times]{sagej}

\usepackage{moreverb, url,multicol,amsmath,subfigure,bm,algorithm,algpseudocode,multirow,empheq,graphicx,amssymb,xcolor,siunitx, soul,booktabs,tabularx,makecell, arydshln}
\usepackage[most]{tcolorbox}
\usepackage[colorlinks,bookmarksopen,bookmarksnumbered,citecolor=blue,urlcolor=blue]{hyperref}
\hypersetup{colorlinks=true, linkcolor=blue, citecolor=blue, urlcolor=blue}
\usepackage{orcidlink}

\newcommand{\trsp}{{\scriptscriptstyle\top}}

\newcommand{\mc}{\mathcal}
\newcommand{\mb}{\mathbb}
\newcommand\BibTeX{{\rmfamily B\kern-.05em \textsc{i\kern-.025em b}\kern-.08em
T\kern-.1667em\lower.7ex\hbox{E}\kern-.125emX}}

\def\volumeyear{2026}
\begin{document}

\runninghead{A. Razmjoo, T. Xue, S. Shetty, and S. Calinon}

\title{Sampling-Based Constrained Motion Planning with Products of Experts}

\author{Amirreza Razmjoo\affilnum{1,2} \orcidlink{0000-0003-3826-6608}, Teng Xue\affilnum{1,2} \orcidlink{0009-0001-7414-3958}, Suhan Shetty\affilnum{1,2} \orcidlink{0000-0002-7550-9368}, and Sylvain Calinon\affilnum{1,2} \orcidlink{0000-0002-9036-6799}}

\affiliation{\affilnum{1}Idiap Research Institute, Martigny, Switzerland\\
\affilnum{2}École Polytechnique Fédérale de Lausanne (EPFL), Lausanne, Switzerland}

\corrauth{Amirreza Razmjoo,\\  Centre du Parc, Rue Marconi 19, CH-1920 Martigny, Switzerland}

\email{amirreza.razmjoofard@epfl.ch}

\begin{abstract}
We present a novel approach to enhance the performance of sampling-based Model Predictive Control (MPC) in constrained optimization by leveraging products of experts. Our methodology divides the main problem into two components: one focused on optimality and the other on feasibility. By combining the solutions from each component, represented as distributions, we apply products of experts to implement a \emph{project-then-sample} strategy. In this strategy, the optimality distribution is projected into the feasible area, allowing for more efficient sampling. This approach contrasts with the traditional \emph{sample-then-project} and na\"ive \emph{sample-then-reject} method, leading to more diverse exploration and reducing the accumulation of samples on the boundaries. We demonstrate an effective implementation of this principle using a tensor train-based distribution model, which is characterized by its non-parametric nature, ease of combination with other distributions at the task level, and straightforward sampling technique. We adapt existing tensor train models to suit this purpose and validate the efficacy of our approach through experiments in various tasks, including obstacle avoidance, non-prehensile manipulation, and tasks involving staying in a restricted volume. Our experimental results demonstrate that the proposed method consistently outperforms known baselines, providing strong empirical support for its effectiveness. Sample codes for this project are available at \href{https://github.com/idiap/smpc_poe}{https://github.com/idiap/smpc\_poe}
\end{abstract}


\maketitle

\section{Introduction}

For a robot to navigate and interact effectively within its environment, it must plan its motion while taking surrounding conditions into account. In most cases, the robot must satisfy various constraints imposed by physics, the environment, or its own mechanical limitations. Adhering to these constraints is often what makes motion planning and control a challenging problem in robotics.

In recent years, optimization-based methods, such as optimal control (OC) \citep{todorov2002optimal} and reinforcement learning (RL) \citep{sutton2018reinforcement}, have led to significant advancements in robotics. These methods are generally categorized into gradient-based and sampling-based approaches. Sampling-based methods, which are gradient-free, are particularly useful in scenarios such as contact-rich manipulation \citep{pezzato2023sampling} and obstacle avoidance \citep{bhardwaj2022storm}, where the system gradient is often non-smooth due to factors such as sudden environmental contact.

\begin{figure}[tbh]
   \centering
    \subfigure[Planar Pushing]{%
        \includegraphics[height=0.2\textwidth]{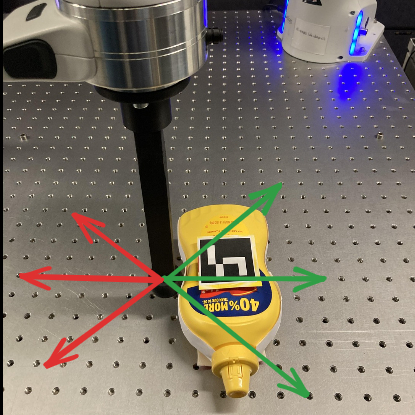}
        }
    \hfill
    \subfigure[Whole-body Obstacle Avoidance]{%
        \includegraphics[height=0.17\textwidth]{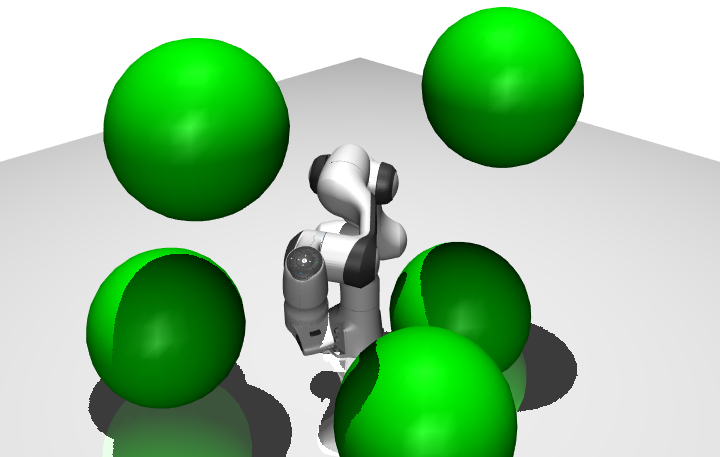}
        }
     \caption{In sampling-based Model Predictive Control (SMPC), many samples can be discarded due to inefficiencies or infeasibility. In (a), only the green samples effectively influence the movement of the object, while the actions indicated by the red arrows have no impact on the environment. (b) illustrates the importance of selecting samples that do not collide with obstacles.}
     \label{fig:efficiency_schematic}
\end{figure} 

These methods have been widely applied in trajectory optimization~\citep{orthey2023sampling}, including well-known techniques like Rapidly-Exploring Random Trees (RRT)~\citep{kuffner2000rrt}. Their gradient-free nature makes them suitable for applications where computing gradients is difficult or infeasible. However, as these methods are only probabilistically complete, they may require extensive sampling. To address this, a common strategy is to reduce the planning horizon and focus on local decision-making. In this context, sampling-based model predictive (SMPC) approaches such as~\citep{williams2017information,bhardwaj2022storm} have been developed to improve responsiveness and adaptability. However, a primary challenge is the lack of efficient sampling mechanisms to capture essential information, particularly in cases involving constraints, sparse cost or reward functions. For instance, Fig. \ref{fig:efficiency_schematic}-a shows a robot trying to push a mustard bottle toward a target, but certain samples can lead it away from the object. Similarly, in Fig. \ref{fig:efficiency_schematic}-b, many samples can collide with the obstacles, resulting in less informative or rejected samples. In this article, we address the challenge of efficient sampling in the presence of constraints by proposing a novel approach to enhance sampling efficiency.

\begin{figure}[tb!]
   \centering
    \subfigure[Sample-then-project]{%
        \includegraphics[width=0.5\textwidth]{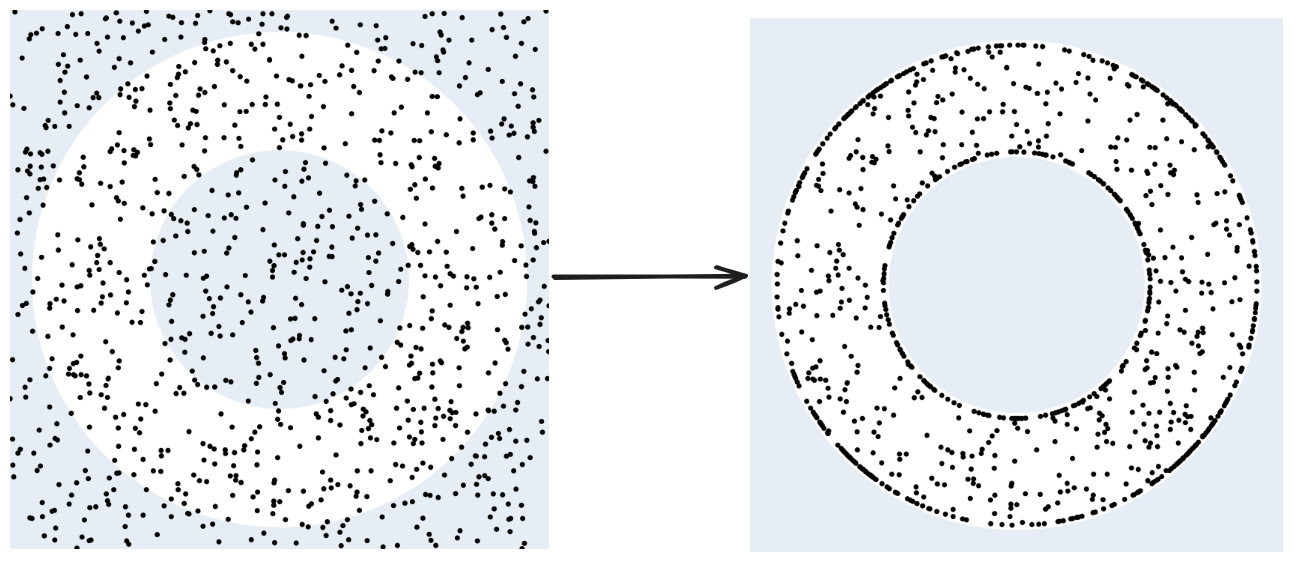}
        }
    \hfill
    \subfigure[Project-then-sample]{%
        \includegraphics[width=0.5\textwidth]{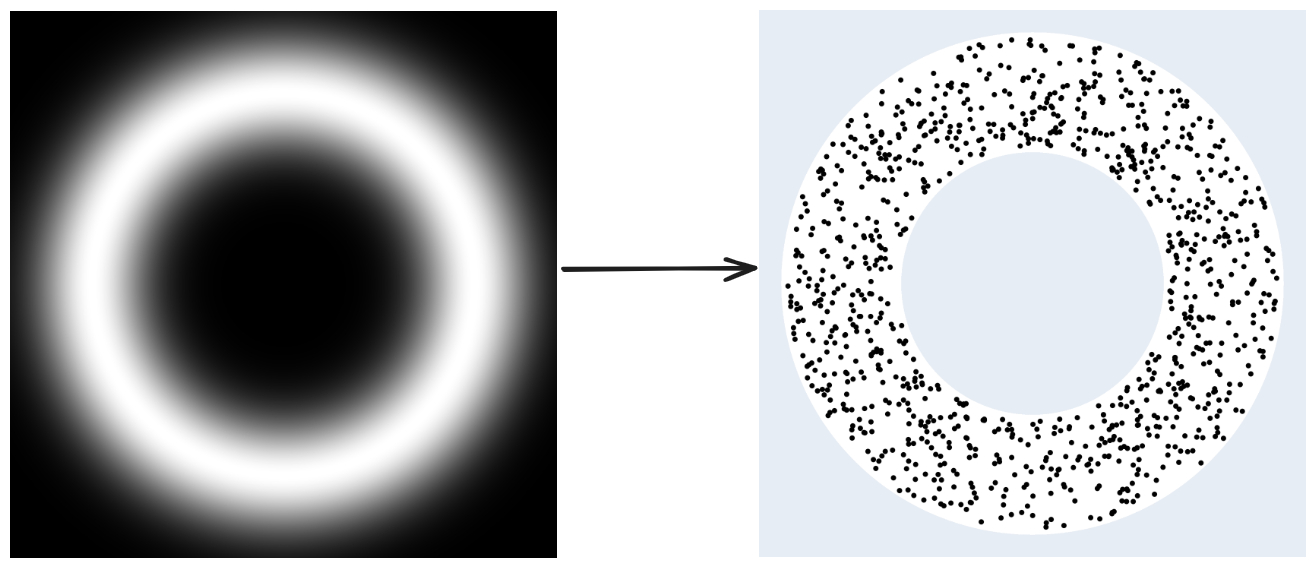}
        }
     \caption{An illustrative example highlighting the difference between two sampling strategies from a uniform distribution that meets the task constraints outlined in Fig. \ref{fig:whole_pipeline}: (a) \emph{sample-then-project} and (b) \emph{project-then-sample}. In (a), many samples cluster on the boundaries, leaving significant areas within the feasible set unexplored, whereas (b) promotes a more thorough exploration. Note that the difference between the two approaches would become more pronounced if the feasible space were narrower.}
     \label{fig:diff_sampling}
\end{figure} 

A basic approach to incorporating constraints into sampling-based methods is to generate samples from a distribution naively without considering constraints and subsequently reject those that violate the constraints. This \emph{sample-then-reject} strategy can be highly inefficient in tightly constrained problems, leading to a large number of discarded samples. This inefficiency has motivated the development of numerous alternative approaches in the literature that improve the sampling process to reduce the number of rejected samples. These methods can be categorized into three main approaches. The first category includes projection-based methods \citep{hpp2016,proj2016}, which sample first and then project into the feasible region. 
However, projection is not always straightforward: some methods \citep{pardo2015projectionbasedbodymotion, suh2011tangent} require iterative computation of system gradients to perform the projection, which can be problematic when gradients are difficult to obtain—such as in non-prehensile manipulation. As shown in Fig. \ref{fig:diff_sampling}, even with an efficient projection mechanism (often difficult to implement), samples can accumulate near the boundary of the feasible region, leading the system to potentially converge toward that boundary. The second approach involves penalizing constraint violations \citep{bhardwaj2022storm}, serving as a data-driven, iterative alternative to projection. However, this method does not guarantee feasible solutions before convergence and may require multiple iterations to achieve feasibility. The third approach, similar to \cite{power2023variational} and \cite{sacks2023learning}, learns a distribution that promotes sampling of feasible and optimal actions or states. While this approach can be effective, it is often complex to implement and computationally expensive (similarly to reinforcement learning methods).

One idea to solve these complexities can be to use several experts. Leveraging multiple expert models to simplify complex tasks is a widely adopted strategy in the machine-learning community. This is commonly implemented either through \emph{mixture of experts} (MoE) \citep{Jordan1993HierarchicalMO}, which combines outputs in a manner analogous to an \emph{OR} operation, or \emph{products of experts} (PoE) \citep{hinton1999poe}, which integrates outputs similarly to an \emph{AND} operation. In our framework, since the objective is to identify solutions that are both optimal \emph{AND} feasible, we adopt the \emph{products of experts} paradigm. Beyond their inherent conceptual differences, PoE offers distinct advantages that make it particularly well-suited to our framework. Specifically, PoE enables each expert to focus on their own decision variables and objectives independently, simplifying and streamlining the learning process for each expert. For instance, in the example presented in this article, the expert tasked with optimizing for action optimality can concentrate solely on action variables (as in methods like MPPI), while another expert can focus on both action and state variables, or even exclusively on state variables, prioritizing feasibility without concern for optimization. By combining their outputs, PoE identifies the intersection of their respective distributions, yielding solutions that satisfy all variables and objectives simultaneously. In contrast, MoE typically combines solutions as a weighted average, potentially failing to account for such considerations. 

Similarly, in other works on SMPC, such as~\citep{pan2024modelbased}, researchers have employed multiple experts to define the overall distribution. In these approaches, each expert specifies a distribution over the full trajectory, and the experts are combined first, either through a summation of cost terms or a product of distributions, before jointly solving the resulting problem. In contrast, our method decomposes the problem differently. The feasibility expert is learned independently and then combined with the optimality distribution. By taking the product of these two distributions, we reduce the probability of the optimality distribution in the infeasible area, thereby improving the efficiency of the sampling process. The optimality distribution can be modeled as a Gaussian distribution using methods such as Model Predictive Path Integral (MPPI), so this part of the problem is computationally efficient and does not require any learning. By introducing a feasibility expert and integrating it with optimality considerations, we obtain the \emph{project-then-sample} strategy, which efficiently focuses the search on the feasible region of the optimization problem, thereby improving overall efficiency.

\begin{figure}[tb!]
   \centering
        \includegraphics[width=0.95\columnwidth]{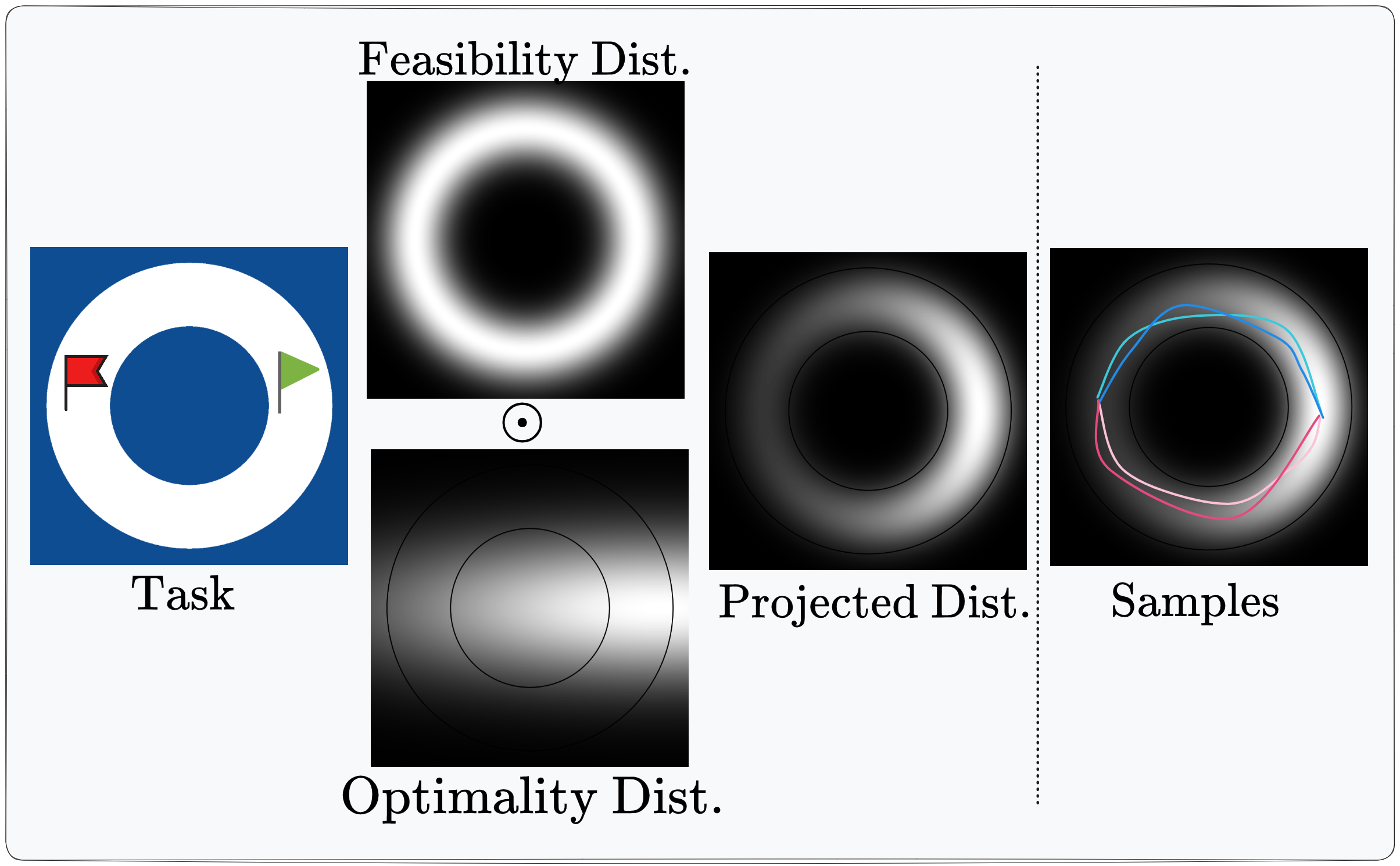}
    \hfill

     \caption{An illustrative example of using products of experts. The agent must navigate from the red flag to the green flag while remaining in the white area and avoiding the blue one. The problem is divided into two parts: one models the feasibility distribution, and the other models the optimality distribution. These are combined to project the optimality distribution into the feasible area (Project first), from which trajectories are then sampled (Then sample).}
     \label{fig:whole_pipeline}
\end{figure}

Fig. \ref{fig:whole_pipeline} illustrates our proposed approach, which splits the main problem into two sub-problems: the \textbf{optimality problem}, focusing on optimization, and the \textbf{feasibility problem}, focusing on constraints. We compute distributions for both subproblems and then combine them with the products of experts approach to create a distribution that is both feasible and optimal. This framework effectively projects the optimality distribution into the feasible region. A key remaining challenge is how to learn these typically complex, multimodal distributions and to sample from them efficiently.

There are three main paradigms for combining experts at the solution level in the literature: Gaussian experts \citep{Calinon16JIST,hinton1999poe}, energy-based models \citep{gkanatsios2023energybased}, and diffusion-based models \citep{yang2023diffusion,wang2024poco}. Gaussian experts offer closed-form inference but are inherently restrictive in the kinds of multimodal distributions they can represent. Energy-based models are more flexible but impose heavy sampling burdens (e.g., Langevin dynamics or MCMC). Diffusion-based approaches significantly reduce sampling complexity, yet typically require large demonstration datasets. Critically, these models often fail to consider feasibility when they are performing sampling during the denoising process, leading to the rejection of many samples at each step. In this paper, we show that by integrating a dedicated feasibility expert, we can improve its reliability before sampling and boost per-step sampling efficiency throughout the diffusion process.

In this article, we demonstrate how different experts can be effectively integrated using tensor train (TT) decomposition. Our previous work demonstrated the potential of this method for learning and sampling from desired distributions \citep{Shetty23,Xue24RSS}. We extend these findings by applying TT in the context of products of experts. We compare it to a baseline relying on normalizing flows (NFs) \citep{dinh2016density}, inspired by \citep{power2023variational}, to show the advantages of TT over NFs, as well as to discuss the differences between a \emph{project-then-sample} strategy compared to a \emph{sample-then-project} or pure \emph{sample-then-reject} strategies. 

In summary, our contributions are threefold:

\begin{itemize}
   \item Enhancing the sampling efficiency of SMPC methods by introducing the feasibility expert and combining it with the optimality distribution using products of experts (i.e., \emph{project-then-sample} strategy).  Particularly, the feasibility distribution is learned separately and then combined with the optimality one before sampling. This enables more effective exploration of the feasible set and extends applicability to problems where gradient calculation is challenging.
    \item Introducing an effective approach to leverage products of experts through TT density estimation, a non-parametric distribution capable of learning multi-modal distributions. This approach can also leverage existing straightforward sampling algorithms, enabling efficient exploration of complex distribution spaces.
    \item Conducting a series of experiments in both simulation and real-world settings to validate the effectiveness of our proposed method compared to baselines.
\end{itemize}

The remainder of this paper comprises a literature review on various aspects related to our work in Section \ref{sec:related} and a concise yet essential background on the employed methods in Section \ref{sec:back}. Our proposed method is detailed in Section \ref{sec:method}, and the results are presented in Section \ref{sec:res}. Finally, we discuss different aspects of the method in Section~\ref{sec:discuss} and conclude the paper in Section \ref{sec:conclusion}.

\section{Related work}
\label{sec:related}
\textbf{Gradient-Based vs Sampling-based MPC:} Analytical solutions are often impractical for optimization problems in robotics, especially with non-convex, non-linear cost functions and constraints arising from kinematic chains and dynamics. Gradient-based methods like iLQR \citep{todorov2002optimal, diehl2006fast} are widely used, offering scalability and local solution updates. However, they face issues such as local minima, and undefined gradients in tasks like contact-rich manipulation \citep{pang2023global,Dafle14ICRA}. These limitations are critical in Model Predictive Control (MPC), where fast solutions are essential.

To address these challenges, sampling-based methods have gained attention. They bypass the need for gradients, can exploit GPU-based simulators for parallel sampling \citep{liang2018gpu}, and may reduce sensitivity to local minima by approximating smoothed gradients \citep{pang2023global}. This makes them promising alternatives for complex robotic tasks involving discrete decisions \citep{dafle2020motioncones}.

\textbf{Sampling-based MPC (SMPC):} Methods such as Rapidly-Exploring Random Tree (RRT) \citep{kuffner2000rrt} and Probabilistic Roadmap Planner (PRM) \citep{kavraki1996probabilistic} for trajectory optimization, as well as approaches such as Cross-Entropy Motion Planning (CEM) \citep{kobilarov2012cross} and Model Predictive Path Integral (MPPI) \citep{williams2017information} have shown the effectiveness of sampling strategies in addressing MPC problems. Recent findings by DeepMind \citep{howell2022predictive} suggest that even for a challenging task such as quadruped walking, selecting the best sample suffices to achieve the objective. However, these advantages are limited to tasks without substantial constraints. 

\textbf{Constrained Sampling-based Motion Planning:} Given that sampling-based methods offer probabilistic completeness, they may require extensive data, posing challenges to finding constraint-free solutions for highly constrained problems. Projection methods are widely used in sampling-based methods, particularly within RRT, to ensure constraints are met and improve sampling efficiency \citep{kingston2018sampling}. Prior research shows that this approach enables probabilistic completeness \citep{berenson2011constrained}, ensuring that the feasible set will be fully explored over time. However, finding an analytical solution for projection remains challenging under most constraints. In specific cases, projection operators can be locally defined using the tangent space and a nullspace projector \citep{pardo2015projectionbasedbodymotion} or by direct sampling within the tangent space \citep{suh2011tangent}. Nevertheless, this approach has limitations, as it only captures local constraint behavior and fails with non-differentiable constraints (e.g., contact-rich manipulation). To address this, sampling-based methods often use gradient-based techniques to approximate the projection \citep{stilman2010global,yakey2001randomized}. Some works, such as \citep{stilman2010global}, frame projection as a secondary optimization problem. An alternative is to relax constraints by introducing penalty terms within the main objective function \citep{bhardwaj2022storm}. However, both approaches face limitations, including increased computational demands and the risk of converging to infeasible regions.

This article contends that the above approaches may not be the most efficient, even with an analytical projection solution. Standard projection-based sampling may guide exploration toward regions with limited feasibility (e.g., boundaries), yielding biased solutions and insufficiently exploring the feasible set. A more effective approach could be to directly model the distribution of the feasible and optimal set and sample exclusively from this distribution, concentrating exploration solely within feasible regions. Recent advances in generative models have facilitated this, allowing feasibility-aware sampling strategies \citep{sacks2023learning,power2023variational}. However, these methods, akin to RL, often require learning highly complex distributions, which can be data-intensive and computationally costly.

Our approach seeks to combine the advantages of both projection and generative methods. We propose a two-stage process: first, we train generative models to approximate the distribution of the feasible region. Rather than projecting individual samples, we then project an estimated optimality distribution into this feasible region using a product of experts, enhancing both the efficiency and focus of sampling within feasible areas. Our approach has two main benefits. Compared to learning a distribution that captures both feasibility and optimality, we focus on the simpler task of feasibility alone and allowing the learned model to be used across tasks with similar feasibility sets. Furthermore, compared to projection-based methods, our approach \emph{project-then-sample} promotes better exploration within the feasible set.

\textbf{Product of Experts:} In SMPC, it is common practice to define the overall cost function or solution distribution by combining multiple terms, typically through summation or a product of distributions~\citep{pan2024modelbased}. Each term is designed to address a specific aspect of the problem, such as optimality or constraint satisfaction, and the final solution is obtained by solving the joint problem defined by their combination.

In most existing approaches, this combination occurs at the definition level, meaning that the full problem is formed first and then solved using techniques such as gradient descent, Langevin dynamics, or other sampling-based methods. While this is effective, it tends to reduce flexibility, as the individual structures of the constituent terms are lost once they are merged.

In contrast, Hinton’s original formulation of the \emph{product of experts}~\citep{hinton1999poe} suggests an alternative: solving each expert individually and then combining their solutions. By \emph{solution}, we refer to a sampleable distribution, such as a Gaussian, that does not require further optimization or inference steps. This approach preserves the modularity of each expert, enables the use of specialized solvers, and simplifies sampling from the final distribution without re-solving the full joint problem. In this regard, each expert represents a relatively simple distribution focusing on specific parameters or tasks. These experts are subsequently combined to yield a more refined and sharper distribution. This methodology has found applications in robotics, including the fusion of sensory information \citep{pradalier2003expressing}, the fusion of robot motion learned in multiple coordinate systems \citep{Calinon16JIST}, and across task spaces \citep{Pignat22IJRR,Muehlbauer24RAL}.  Our work extends these prior approaches by incorporating the product of experts framework within an MPC paradigm. 

There are three main approaches in the literature for combining multiple experts at the solution level: (1) \textbf{Gaussian experts}~\citep{Calinon16JIST, hinton1999poe}, (2) \textbf{Energy-based models}~\citep{gkanatsios2023energybased}, and (3) \textbf{Diffusion-based models}~\citep{yang2023diffusion, wang2024poco}. Each of these methods offers different advantages and trade-offs, which we discuss further to motivate our use of the TT representation.

Compared to these methods, Gaussian distributions are significantly more limited than TT and other baselines, primarily due to the parametric nature of Gaussian models. Energy-based models \citep{lecun2006tutorial}, which can be viewed as a more general form of Gaussian distributions, indeed hold greater potential. However, according to our experience, sampling from these models typically relies on Langevin dynamics or methods such as MCMC, which can become a significant computational bottleneck. In contrast, TT does not suffer from this issue and allows for efficient and straightforward sampling.

Diffusion models \citep{ho2020denoising} (or similar variations such as Flow-Matching \citep{lipman2023flow}) address the sampling challenges present in energy-based models to a significant extent. These models can be broadly categorized into two types: data-based~\citep{wang2024poco} and model-based~\citep{pan2024modelbased}. Data-based methods require the optimality distribution to be explicitly learned in diffusion form, which typically demands a substantial number of high-quality demonstrations—a requirement not assumed in our work. 

Moreover, recent works such as~\citep{xue2025icra, pan2024modelbased} have proposed innovative strategies to represent diffusion models in a data-free manner. Their approach, which bears resemblance to annealed MPPI methods, combines multiple experts at the function level. In their pipeline, different candidate solutions are sampled first, followed by a rejection step to discard infeasible samples at each denoising step. By contrast, our approach merges experts at the solution level. This allows us to reshape the sampling distribution before sampling, which is a more proactive method than relying solely on post-hoc sample rejection. Concretely, we inject the feasibility expert’s density into every denoising (sampling) step of the diffusion process, producing candidates that are inherently more likely to satisfy constraints.

\section{Background}
\label{sec:back}
\subsection{Model Predictive Path Integral}\label{subsec:MPPI}
The primary aim of this methodology is to tackle an optimal control problem associated with a dynamic system described by $\bm{x}_{t+1} = f_{\text{dyn}}(\bm{x}_t, \bm{u}_t, \bm{\omega}_t)$ within a receding horizon $H$. Here, $\bm{x}_t \in \mb{R}^{d_x} $ and $\bm{u}_t \in \mb{R}^{d_u}$ denote the system state and action at time $t$, respectively, while $\bm{\omega}_t$ represents a stochastic noise element in the system. This objective is achieved by minimizing a predetermined cost function as
\begin{equation}
    c(\hat{\bm{X}}_t,\hat{\bm{U}}_t) = \sum_{h=0}^{H-1} c_h(\hat{\bm{x}}_{t+h},\hat{\bm{u}}_{t+h}) + c_H(\hat{\bm{x}}_{t+H}).
\end{equation}

Here, $\hat{\bm{X}}_t = [\hat{\bm{x}}_t^\trsp, \hat{\bm{x}}_{t+1}^\trsp, \ldots, \hat{\bm{x}}_{t+H}^\trsp]^\trsp$ represents the state trajectory, and $\hat{\bm{U}}_t = [\hat{\bm{u}}_t^\trsp, \hat{\bm{u}}_{t+1}^\trsp, \ldots, \hat{\bm{u}}_{t+H-1}^\trsp]^\trsp$ denotes the sequence of action commands. The hat symbol \(\hat{(\cdot)}\) indicates that these actions are part of the planned, but not yet applied, control sequence. Additionally, $c_h$ and $c_H$ denote stage and terminal cost values, respectively. Here, the constraints are formulated inside the cost function. The objective is to determine a set of optimal action commands $\bm{U}_t^*$ that minimizes this cost function while taking into account the dynamics of the system, specifically
\begin{equation}
    \bm{U}^*_t = \underset{\hat{\bm{U}}_t}{\text{argmin}}  \ c(\hat{\bm{X}}_t,\hat{\bm{U}}_t).
\end{equation}

Williams et al. \citep{williams2017model, williams2017information} demonstrated that the solution for this optimal control problem can be effectively approximated using a forward approach, diverging from the conventional dynamic programming method. This discovery opened avenues for leveraging Monte Carlo estimation in approximating optimal actions through sampling-based methods. Furthermore, \cite{wagener2019online} showed that if the system motion can be modeled as a Gaussian distribution such as
\begin{equation}
    \pi_{\theta}(\hat{\bm{U}}_t) = \prod_{h=0}^{H-1}\pi_{\theta_h}(\hat{\bm{u}}_{t+h}) = \prod_{h=0}^{H-1}\mc{N}(\hat{\bm{u}}_{t+h}|\bm{\mu}_{t+h},\bm{\Sigma}_{t+h}),
\end{equation}
then its parameters can be approximated as 
\begin{equation}\label{eq:mppi_update}
    \bm{\mu}_{t} = (1-\gamma_t) \hat{\bm{\mu}}_{t} + \gamma_t \sum_{i=1}^{N} w_i \hat{\bm{U}}^{(i)}_{t}, 
\end{equation}
where $N$ denotes the number of samples, $\hat{\bm{\mu}}_{t}$ represents the preceding mean value, and $\gamma_t$ is the step size. The term $w_i$ is a weighting factor expressed as
\begin{equation}\label{eq:mppi_weights}
     w_i = \frac{e^{-\frac{1}{\beta} c(\hat{\bm{X}}^{(i)}_{t},\hat{\bm{U}}^{(i)}_{t})}}{\sum_{j=0}^{N}e^{-\frac{1}{\beta} c(\hat{\bm{X}}^{(j)}_{t},\hat{\bm{U}}^{(j)}_{t})}},
\end{equation}
where $\beta$ is a temperature parameter. Following the calculation of the parameters, the initial action $\bm{u}^*_t$ can be sampled, applied to the system, and iteratively repeated until the task is successfully completed.

\subsection{Tensor Train Decomposition}\label{sec:TT_dist}
A $d$-th order tensor, denoted as $\bm{\mc{P}} \in \mb{R}^{n_{1}\times\cdots \times n_{d}}$, serves as an extension of vectors (1-D) and matrices (2-D) to higher dimensions ($d$-D). Its configuration is defined by a tuple of integers $\bm{n} = (n_1, \ldots, n_d)$. Each element in the tensor, denoted as $\bm{\mc{P}}_{\bm{i}}$, is identified by $\bm{i} = (i_1, \ldots, i_d)$ as a set of $d$ integer variables representing the index in each dimension. The index set of the tensor is expressed as $\mc{I} = \{{\bm{i} = (i_1, \ldots, i_d): i_k \in \{1, \ldots, n_k\}, k \in \{1, \ldots, d\}}\}$.

We can express a function, denoted as $p: \Omega_{\bm{x}} \subset \mb{R}^d \rightarrow \mb{R}$, as a tensor by assuming a rectangular domain for the function, such as $\Omega_{\bm{x}} = \Omega_{x_1}\times\Omega_{x_2}\times\cdots\times\Omega_{x_d}=\times_{k=1}^d \Omega_{x_k}$. This involves discretizing each input interval $\Omega_{x_k} \subset \mb{R}$ into $n_k$ elements. The continuous set $\mc{X} = \{ \bm{x} = (x^{i_1}, \ldots, x^{i_d}): x^{i_k} \in \Omega_{x_k}, i_k \in {1, \ldots, n_k} \}$ can be easily mapped to the corresponding index set $\mc{I}_{\mc{X}} = \{ \bm{i} = (i_1, \ldots, i_d) \}$. The $\bm{i}$-th element in the tensor $\bm{\mc{P}}$, denoted by $\bm{\mc{P}}_{\bm{i}}$, corresponds to the function value with continuous variables, such that $\bm{\mc{P}}_{\bm{i}} = p(\bm{x})$.  We overload the terminology and define  $\bm{\mc{P}}_{\bm{x}} = \bm{\mc{P}}_{\bm{i}}$ where $\bm{i}$ is the corresponding index of $\bm{x}$. Interpolation techniques between specific nodes of the tensor $\bm{\mc{P}}$ can be used to approximate the value $p(\bm{x})$ for any $\bm{x} \in \Omega_{\bm{x}}$.

\begin{figure}[tb!]
    \centering
     \includegraphics[width=0.9\linewidth]{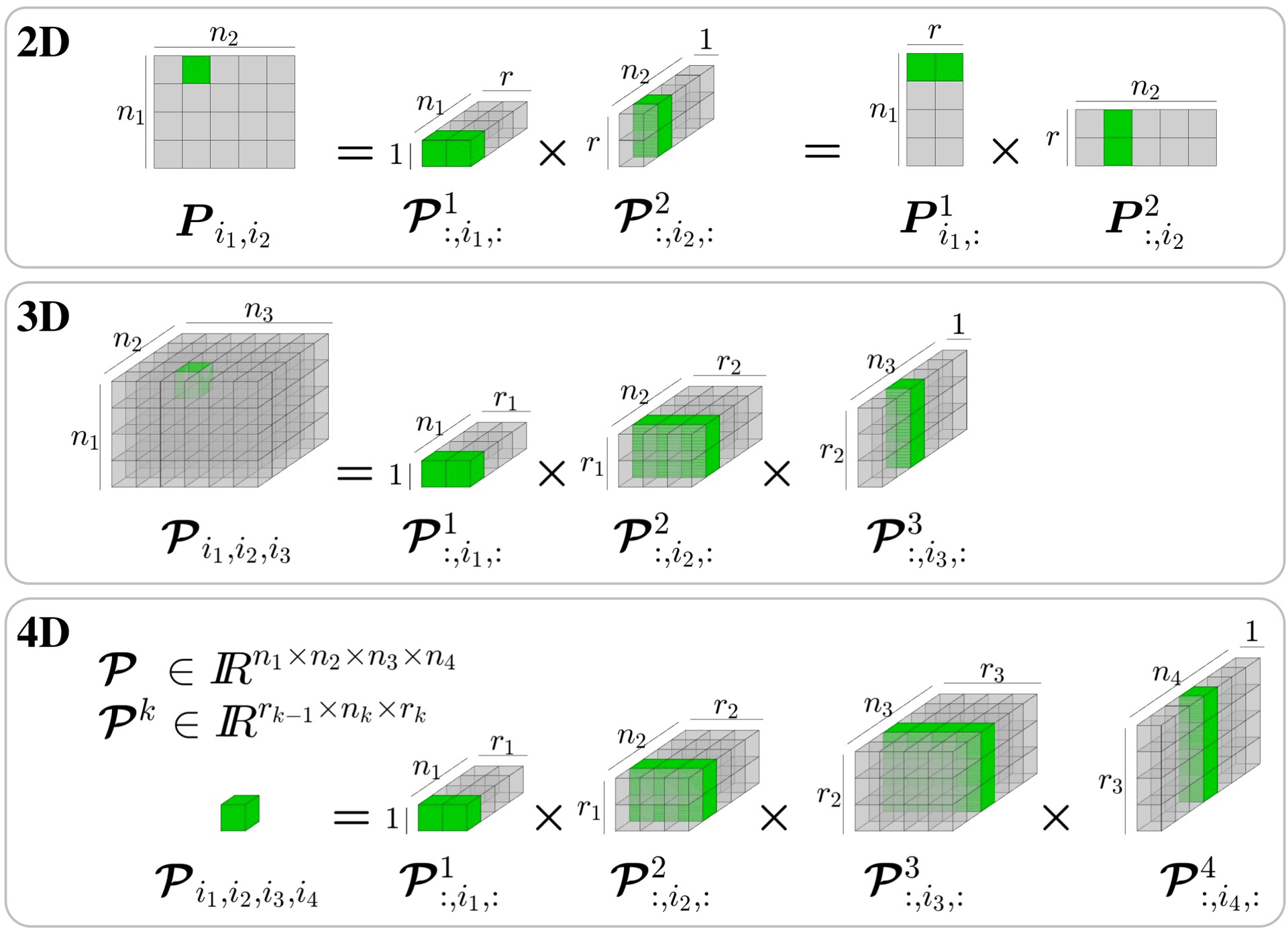}
     \caption{TT decomposition extends matrix decomposition techniques to higher-dimensional arrays. In the TT format, accessing an element of a tensor involves multiplying the chosen slices (represented by green-colored matrices) of the core tensors (factors). The illustration provides examples for a 2nd order, 3rd order, and a 4th order tensor. This picture is adopted from \cite{Shetty23} with permission from the authors.}
     \label{fig:tt_format}
\end{figure} 

To represent a tensor efficiently for a high-dimensional function, we employ a tensor factorization method that enhances storage efficiency by utilizing factors with a reduced number of elements. Among various methodologies, we adopt the Tensor Train (TT) decomposition approach in this study, which utilizes a set of third-order tensors known as \textit{TT-cores} to compactly encode a given tensor. In the TT format, a $d$-th order tensor $\bm{\mc{P}} \in \mb{R}^{n_1\times\cdots \times n_d}$ is expressed using a tuple of $d$ third-order tensors $(\bm{\mc{P}}^1,\ldots,\bm{\mc{P}}^d)$. The dimensions of the TT-cores are denoted as $\bm{\mc{P}}^k\in \mb{R}^{r_{k-1} \times n_k \times r_{k} }$ with $r_0 = r_d = 1$. As illustrated in Fig.~\ref{fig:tt_format}, the computation of the $\bm{i}$-th element of the tensor in this format involves a straightforward multiplication of matrix slices obtained from the TT-cores as
\begin{equation}
    \label{eq:tt_rep}
    \bm{\mc{P}_i} = \bm{\mc{P}}^1_{:,i_1,:}\bm{\mc{P}}^2_{:,i_2,:}\cdot\cdot\cdot \bm{\mc{P}}^d_{:,i_d,:},
\end{equation}
where $\bm{\mc{P}}^k_{:,i_k,:} \in \mb{R}^{r_{k-1} \times r_k}$ denotes the $i_{k}$-th matrix slice of the third-order tensor $\bm{\mc{P}}^k$. The multiplication of these TT-cores yields a scalar variable. The determination of these cores, which have been demonstrated to exist for any given tensor consistently \citep{oseledets2011tensor}, can be achieved through various methods, such as TT-SVD \citep{oseledets2011tensor} and TT-cross \citep{savostyanov2011fast}.

The \textit{TT-rank} of the tensor in the TT representation is defined by the tuple $\bm{r} = (r_{1}, r_{2}, \ldots, r_{d-1})$. We refer to $r = \max{(r_1, \ldots, r_{d-1})}$ as the maximal rank. Similar to the matrix analogy, a smaller $r$ leads to a less precise approximation, while a larger $r$ necessitates greater storage capacity.

Utilizing this methodology, the ultimate outcome is the approximation of the continuous function $p$, expressed as
\begin{equation}
    \label{eq:seperability_matrix}
    p(x_1,\ldots,x_d) \approx \bm{P}^1(x_1) \cdots \bm{P}^d(x_d).
\end{equation}

Here, $\bm{P}^k(x_k)$ with $k \in \{1, \dots, d\}$ is derived by interpolating within each of the cores. Employing advanced interpolation techniques enables a coarser discretization, thereby reducing storage requirements during the learning phase.

\subsection{Tensor-Train Distribution}
The TT approximation technique has been previously applied to optimization problems  \citep{batsheva2023protes, Shetty23}. In this approach, the problem of minimizing a cost function is first transformed into the task of maximizing an unnormalized probability density function $p(\cdot)$, represented as a tensor $\bm{\mc{P}}$ in TT format. Subsequently, a corresponding probability distribution, referred to as the TT distribution, is constructed as
\begin{equation}
    \label{eq:tt_distribution}
    \text{Pr}(\bm{x}) = \frac{|\bm{\mc{P}}_{\bm{x}}|}{Z}, \quad \bm{x} \in \Omega_{\bm{x}},
\end{equation}
where $Z$ denotes the corresponding normalization constant. Leveraging the separable structure of the TT model allows for the efficient generation of exact samples from the TT distribution, eliminating the need to compute the normalization factor $Z$. Additional details about the sampling process can be found in the Appendix. Moreover, as illustrated in \citep{Shetty23}, direct calculations of marginal and conditional distributions can be derived from the TT-cores. 

For simplicity in notation, we employ $\bm{\mc{P}}(\bm{x})$ to denote the tensor element corresponding to $\bm{x}$ and $\bm{\mc{P}}(\bm{y_2}|\bm{y_1})$ to express the conditional probability of $\bm{y}_2 = [x_{j+1}, \ldots, x_{d}]$ given $\bm{y}_1 = [x_1, x_2, \ldots, x_j]$. It is important to emphasize that the latter is directly computed from $\bm{\mc{P}}$.

\section{Methodology}
\label{sec:method}
\subsection{Problem Definition}
In a constrained SMPC problem, the aim is to find an optimal sequence of actions $\hat{\bm{U}}_t = {[\hat{\bm{u}}_t^\trsp, \hat{\bm{u}}_{t+1}^\trsp, \ldots, \hat{\bm{u}}_{t+H-1}^\trsp]}^\trsp$ at time step $t$ over a horizon $H$. This sequence should optimize a cost function as
\begin{equation}
\begin{gathered}\label{eq:cost_function}
    \bm{U}^*_t = \underset{\hat{\bm{U}}_t}{\text{argmin}} \sum_{h=0}^{H-1} c_h(\hat{\bm{x}}_{t+h},\hat{\bm{u}}_{t+h},{\bm{\theta}}) + c_H(\hat{\bm{x}}_{t+H}, {\bm{\theta}}), \\
    \text{s.t. }\bm{f}(\hat{\bm{x}}_{t+h+1},\hat{\bm{u}}_{t+h}, {\bm{\theta}}) \leq \bm{0}, \quad\forall h \in \{0,1,\hdots,H-1\}.
\end{gathered}
\end{equation}

Here, \(\bm{x}_t\) denotes the system state at time step \(t\), \(\bm{f}(\cdot) \in \mathbb{R}^w\) represents the vector of \(w\) constraints, and \(c_h(\cdot)\) and \(c_H(\cdot)\) denote the stage and terminal costs, respectively. $\bm{\theta}$ denotes task-dependent terms such as obstacle locations, target positions, etc. For clarity, we omit these terms in the remainder of the paper; however, they are still incorporated in our experiments, as detailed in the Appendix. It is important to note that the constraint set may incorporate prior knowledge that we expect the system to follow. For instance, in a pushing task, one constraint could ensure that actions taken at each step influence the object and prevent the robot from moving away from it beyond a certain threshold, as illustrated in Fig. \ref{fig:efficiency_schematic}-a.

\subsection{Products of Experts}\label{subsec:poes}
Similarly to other SMPC methods, we assume that the solution to this problem can be modeled as a distribution \(p(\hat{\bm{U}}_t|\bm{x}_t)\). Furthermore, we assume that this distribution can be expressed as a product of two other distributions
\begin{equation}\label{eq:prod_exp_stat_dyn}
    p(\hat{\bm{U}}_t|\bm{x}_t) = p^o(\hat{\bm{U}}_t|\bm{x}_t) p^f(\hat{\bm{U}}_t|\bm{x}_t).
\end{equation}
The first distribution, \(p^o\), captures the dynamic aspect of the task and ensures optimality (i.e., the optimality distribution, Expert 1), while the second distribution, \(p^f\), represents the feasible area (i.e., the feasibility distribution, Expert 2).

\textbf{\textit{Expert 1}}: We can solve  \eqref{eq:cost_function} using the MPPI method by incorporating the constraint within the cost function. This approach enables the system to gradually converge to the feasible area, although it may lead to time-consuming iterations and the rejection of numerous samples due to high-cost values. This expert addresses the following problem:
\begin{equation}
\begin{gathered}\label{eq:cost_function_dynamic}
    \bm{U}^*_t = \underset{\hat{\bm{U}}_t}{\text{argmin}} \sum_{h=0}^{H-1} c_h(\hat{\bm{x}}_{t+h},\hat{\bm{u}}_{t+h}) + \\ c_f(\bm{x}_{t+h+1},\bm{u}_{t+h}) + c_H(\hat{\bm{x}}_{t+H}),
\end{gathered}
\end{equation}
where \( c_f(\cdot) \) denotes the cost term associated with constraint violations in \( \bm{f} \). Although Expert 1 is primarily responsible for modeling the optimality distribution, the inclusion of the additional cost term \(c_f(\cdot)\) offers several advantages. First, by assigning significantly higher values to this term relative to others, infeasible samples are effectively assigned zero weight via \eqref{eq:mppi_weights}, thereby removing them from consideration. Secondly, it allows us to express the objective in a more general form, consistent with how similar formulations are structured in related literature, thereby making Expert 1 directly comparable to prior approaches. Third, this added term provides flexibility to account for certain constraints that may emerge during task execution but are not explicitly captured by Expert 2 (described later). Finally, from a practical standpoint, incorporating \(c_f(\cdot)\) into Expert 1 can enhance efficiency by guiding the optimization using relevant constraint information, even when a separate feasibility expert is also present. In this paper, we follow prior works \citep{Jankowski23ICRA, bhardwaj2022storm, power2023variational} by assigning a large weight to this term in order to more effectively reject infeasible solutions. \cite{Bhardwaj2021ManipulationMPCExperiments} provide a detailed analysis of how increasing this penalty improves feasibility under tight constraints.

The solution to  \eqref{eq:cost_function_dynamic} using the MPPI method can be formulated as
\begin{equation}\label{eq:dyn_dist}
 p^o(\hat{\bm{U}}_t) = \prod_{h=0}^{H-1}\mathcal{N}(\hat{\bm{u}}_{t+h}|\bm{\mu}_{t+h},\bm{\Sigma}_{t+h}),
\end{equation} 
which represents a Gaussian distribution at each time step, characterized by the mean \(\bm{\mu}_{t+h}\) and covariance matrix \(\bm{\Sigma}_{t+h}\).

\textbf{\textit{Expert 2}}: Handling constraints exclusively through a rejection mechanism, for example by assigning high values to the $c_f$ term, is generally inefficient. It wastes computational resources by generating and then discarding infeasible trajectories. Our proposed method addresses this by introducing a \emph{feasibility expert} that proactively biases the sampling distribution toward feasible regions. This improves the quality of samples before they are evaluated and potentially rejected, leading to several key benefits as discussed in this paper.

To modify the sampling distribution, we introduce another distribution \(p^f(\cdot)\) that focuses on the feasible set \(\mathcal{F}: \{(\hat{\bm{x}}, \hat{\bm{u}} \,|\, \bm{f}(\bm{x}, \bm{u}) \leq 0 \}\). We define a function \(p_{\text{feas}}\) to capture this feasible area as 
\begin{equation}\label{eq:const_static}
p^f(\hat{\bm{x}}_t,\hat{\bm{u}}_t) \propto p_{\text{feas}}(\hat{\bm{x}}_t,\hat{\bm{u}}_t)= \exp\Big(-\frac{\text{ReLU}\Big(\bm{f}(\hat{\bm{x}}_t,\hat{\bm{u}}_t)\Big)}{\lambda}\Big), 
\end{equation}
where $\lambda$ is a temperature parameter. Various methods, including different generative models, can be employed to convert this function into a proper distribution. This article chooses the approach outlined in \citep{Shetty23}. This method discretizes the function \eqref{eq:const_static} and transforms it into an unnormalized distribution. The resulting distribution can be represented using tensor cores (as defined in Sec. \ref{sec:TT_dist}), enabling a lower memory footprint and seamless integration with the other experts, along with a straightforward sampling method from the new distribution. 

After converting the constraint function \eqref{eq:const_static} into a distribution in TT format, we can express it as
\begin{equation}\label{eq:function_to_tt}
    p^f(\hat{\bm{x}}_t, \hat{\bm{u}}_t) \rightarrow \bm{\mc{P}}_f(\hat{\bm{u}}_{t},\hat{\bm{x}}_{t}) = \prod_{n=0}^{d_x} \bm{\mathcal{P}}^n_{{:,i_{\hat{x}_n},:}}\prod_{m=0}^{d_u} \bm{\mathcal{P}}^{d_x + m}_{{:,i_{\hat{u}_m},:}},
\end{equation}
where $\hat{x}_n$ and $\hat{u}_m$ are respectively $n$-th and $m$-th value of the state $\hat{\bm{x}}_t$ and action $\hat{\bm{u}}_t$ vectors. As discussed in \citep{Shetty23}, it is possible to derive the conditional distribution in TT format \(\bm{\mc{P}}_f(\hat{\bm{u}}_{t+h}|\hat{\bm{x}}_{t+h})\) from the joint distribution obtained in \eqref{eq:function_to_tt}. Consequently, assuming that this distribution is time-independent (though it can be readily extended to a time-dependent version by introducing time as an additional variable into the system), we can define the second expert as

\begin{equation}\label{eq:static_dist}
    p^f(\hat{\bm{U}}_t|\bm{x}_t) \propto \prod_{h=0}^{H-1} \bm{\mc{P}}_f(\hat{\bm{u}}_{t+h}|\hat{\bm{x}}_{t+h}).
\end{equation}

\subsection{TT-PoE-MPPI}\label{subsec:tt_poemppi}
By substituting the values derived in \eqref{eq:static_dist} and \eqref{eq:dyn_dist} into \eqref{eq:prod_exp_stat_dyn}, we can express the distribution as
\begin{equation}
    p(\hat{\bm{U}}_t|\bm{x}_t) \propto \prod_{h = 0}^{H-1} \mathcal{N}(\hat{\bm{u}}_{t+h}|\bm{\mu}_{t+h},\bm{\Sigma}_{t+h}) \bm{\mc{P}}_f(\hat{\bm{u}}_{t+h}|\hat{\bm{x}}_{t+h}).
\end{equation}

Next, we focus on a specific time step \(h\), where the sampling can be performed from the distribution
\begin{equation}
    p^h(\hat{\bm{u}}_{t+h}|\bm{x}_t) \propto \mathcal{N}(\hat{\bm{u}}_{t+h}|\bm{\mu}_{t+h},\bm{\Sigma}_{t+h}) \bm{\mc{P}}_f(\hat{\bm{u}}_{t+h}|\hat{\bm{x}}_{t+h}).
\end{equation}

We can rewrite the conditional distribution as
\begin{multline}
   p^h(\hat{\bm{u}}_{t+h}|\hat{\bm{x}}_{t+h}) \propto \\ \mathcal{N}(\hat{\bm{u}}_{t+h}|\bm{\mu}_{t+h},\bm{\Sigma}_{t+h})  \bm{\mc{P}}_f(\hat{\bm{u}}_{t+h},\hat{\bm{x}}_{t+h}).
\end{multline}

This product can be computed efficiently by introducing a new tensor, \(\bm{\mathcal{P}}_{\mathcal{N}}(\hat{\bm{u}}_{t+h})\), which represents the tensor form of the \(\mathcal{N}(\cdot)\) function. Consequently, the product can be expressed as
\begin{equation}\label{eq:prod_tensor}
    p^h(\hat{\bm{u}}_{t+h}|\hat{\bm{x}}_{t+h}) \propto \bm{\mathcal{P}}_{\mathcal{N}}(\hat{\bm{u}}_{t+h}) \bm{\mc{P}}_f(\hat{\bm{u}}_{t+h},\hat{\bm{x}}_{t+h}).
\end{equation}

Although this product can be computed for any distribution, if we assume the covariance matrix is diagonal (i.e., independent actions in different dimensions), \(\bm{\Sigma}_{t+h} = \text{diag}(\bm{\sigma}^2_{t+h})\), the computation becomes more efficient, allowing it to be performed directly at the core level. This assumption is common \citep{power2023variational,sacks2023learning} and is crucial as it enables efficient multiplication of these distributions.

\begin{figure*}[tb!]
    \centering
     \includegraphics[width=0.8\linewidth]{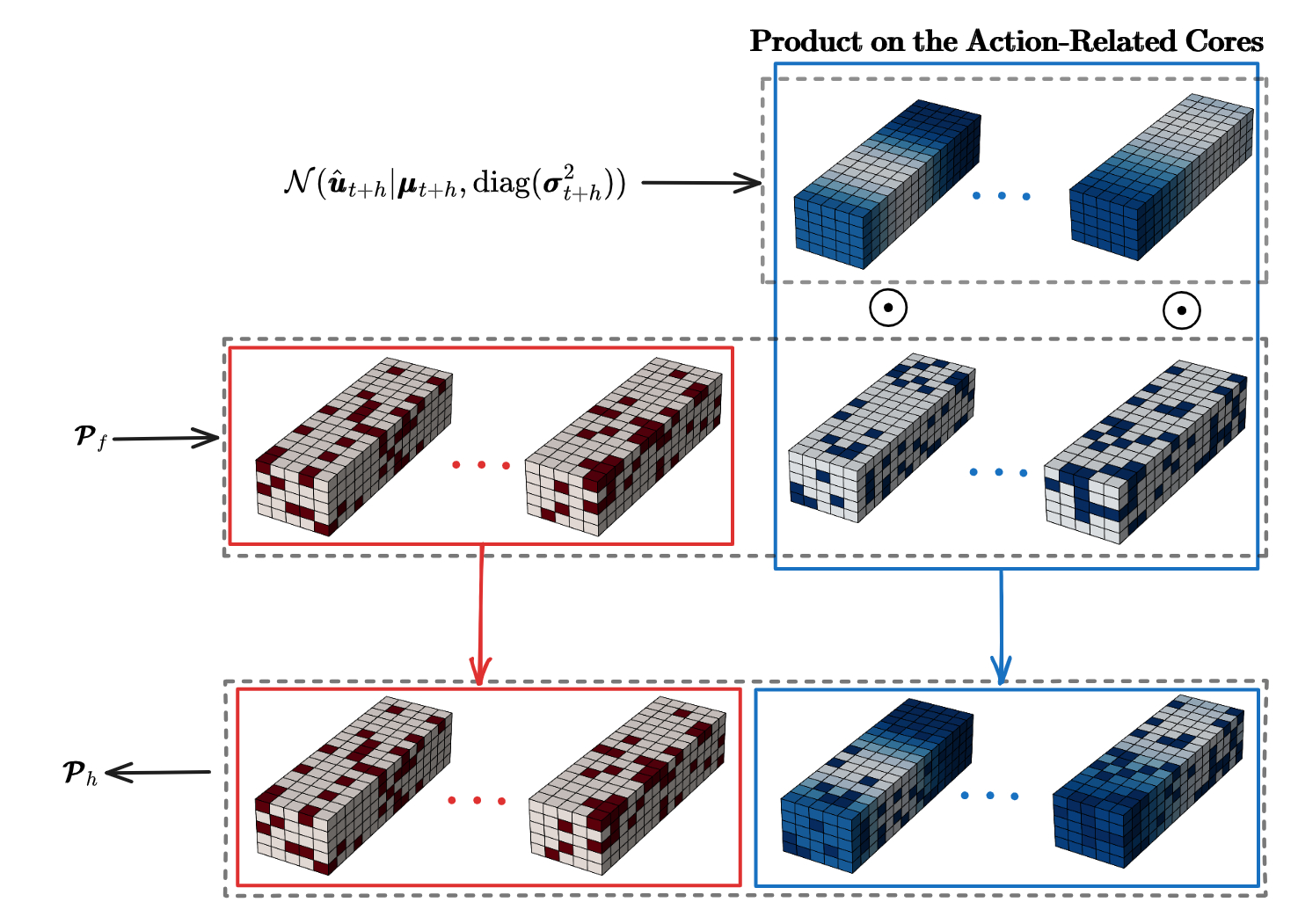}
     \caption{The pipeline employed in this paper integrates two distributions. Red, and blue colors indicate the cores related to the state and action variables, respectively. Initially, the optimality distribution is transformed into TT-cores, followed by elementwise multiplication with the feasibility distribution cores corresponding to the action space. Subsequently, a new distribution is generated using the updated cores. In the illustration, white boxes represent elements equal to 1, while darker boxes denote elements equal to 0. It is important to note that this visualization is simplified for clarity: in actual scenarios, the cores would not strictly be 0 or 1.}
     \label{fig:TTMPPI_schematic}
\end{figure*}

It is important to note that the cost value is calculated via \eqref{eq:cost_function}, and thus the sample weights in \eqref{eq:mppi_weights} are computed by considering the actions across all times and dimensions collectively. Consequently, their influence on one another is captured in the desired action command. However, the independence assumption simplifies the learning and sampling process. Given this assumption, the elementwise multiplication can be independently performed at each TT-core as 
\begin{multline}\label{eq:prod_cores}
    p^h(\hat{\bm{u}}_{t+h}|\hat{\bm{x}}_{t+h}) \propto \bm{\mc{P}}_h \approx \\ \prod_{n=0}^{d_x} \bm{\mathcal{P}}^n_{:,i_{\hat{x}_n},:}\prod_{m=0}^{d_u}{\Big( \bm{\mathcal{P}}^{d_x + m} \odot \bm{\mathcal{P}}_{\mathcal{N}_m}\Big)}_{:,i_{\hat{u}_m},:},
\end{multline}
where \(\odot\) denotes elementwise multiplication, and \(\bm{\mathcal{P}}_{\mathcal{N}_m}\) has the same dimension as \(\bm{\mathcal{P}}^{d_x+m}\), and its elements can be described by
\begin{equation}\label{eq:calc_gauss_cores}
\bm{\mathcal{P}}_{{\mathcal{N}_m}_{:,i_{\hat{u}_m},:}} = \frac{1}{\sigma^m_{t+h}\sqrt{2\pi}} \exp\left(-\frac{1}{2}\left(\frac{\hat{u}_m - \mu^m_{t+h}}{\sigma^m_{t+h}}\right)^2\right),
\end{equation}
where \(\mu^m_{t+h}\) and \(\sigma^m_{t+h}\) represent the values of \(\bm{\mu}_{t+h}\) and \(\bm{\sigma}_{t+h}\), respectively, corresponding to the \(m\)-th dimension. The schematic process for this step at time \(t+h\) is presented in Fig. \ref{fig:TTMPPI_schematic}. 

Once we determine the desired distribution, we employ the same procedure as in SMPC methods. We sample a set of actions, evaluate their associated cost values through a forward path, update the mean values accordingly, and iterate through this process until the task is completed. The complete procedure is outlined in Algorithm \ref{alg:process}.

\begin{algorithm}[t]
\caption{TT-PoE-MPPI}\label{alg:process}
\begin{algorithmic}[1]
\Statex \hspace*{-\algorithmicindent} \textbf{Require:} $\bm{x}_t, \bm{\mu}_{t, \ldots,t + H-1}, \bm{\sigma}_{t, \ldots,t+H-1}$
\Statex \hspace*{-\algorithmicindent} \textbf{Return:} $\bm{u}^*_t$
\State $\hat{\bm{x}}_{t} \gets \bm{x}_{t}$
\For{$h$ from $0$ to $H-1$}
\State Calculate $\bm{\mathcal{P}}_{\mathcal{N}_m}$ \Comment{Eq. \eqref{eq:calc_gauss_cores}}
\State $\bm{\mc{P}}_h \gets$Product of the distributions \Comment{Eq. \eqref{eq:prod_cores}}
\State $\hat{\bm{u}}^0_{t+h},\ldots, \hat{\bm{u}}^N_{t+h} \gets $Sample $N$ actions 
\State $\hat{\bm{x}}^0_{t+h+1},\ldots, \hat{\bm{x}}^N_{t+h+1} \gets $Forward Dynamics
\EndFor
\State $c^0, \dots, c^N \gets $Calculate cost values
\State $\bm{\mu}_{t} \gets $ update the mean value \Comment{Eq. \eqref{eq:mppi_update}} 
\State $\bm{u}^*_t \gets \bm{\mu}_t$
\end{algorithmic}
\end{algorithm}

\subsection{Decomposable feasible distribution}
The same idea can be applied to facilitate the learning of more complex feasibility distributions from simpler ones. For instance, without loss of generality, consider a scenario where the feasible area can be defined by two constraints, \(f_1(\bm{x}_t,\bm{u}_t)\) and \(f_2(\bm{x}_t,\bm{u}_t)\). Instead of learning the feasible set for both constraints simultaneously, one could learn them separately as \(\bm{\mathcal{P}}^{f_1}\) and \(\bm{\mathcal{P}}^{f_2}\). This approach not only simplifies the problem but also increases efficiency. In some parts of the task, we may only need to consider one of the constraints, eliminating the need to relearn both.

The feasible set for the entire problem can be easily computed by combining the two separate (TT) models as 
\begin{equation}
    \bm{\mathcal{P}}^f = \bm{\mathcal{P}}^{f_1} * \bm{\mathcal{P}}^{f_2},
\end{equation}
where the symbol \( * \) denotes the product of the two TT models. To calculate this product, we do not need access to the entire tensor, as it can be efficiently computed at the core level. Libraries such as tntorch \citep{usvyatsov2022tntorchtensornetworklearning} can facilitate this process.

As a side note, when the feasible regions of two constraints overlap, we can use a relaxed combination via summation rather than multiplication as \begin{equation}
\bm{\mathcal{P}}^f = \bm{\mathcal{P}}^f_1 + \bm{\mathcal{P}}^f_2,
\end{equation}
where \(+\) denotes arithmetic operations between two TT models. While a product-of-experts formulation is often preferred, the additive combination retains the same high-probability overlap of feasible regions and thus remains effective at guiding the sampler toward the feasible set. Moreover, this relaxed \emph{sum-of-experts} requires fewer ranks and can be more practical in high-dimensional settings.

\begin{figure*}[tb!]
    \centering
     \includegraphics[width=0.95\linewidth]{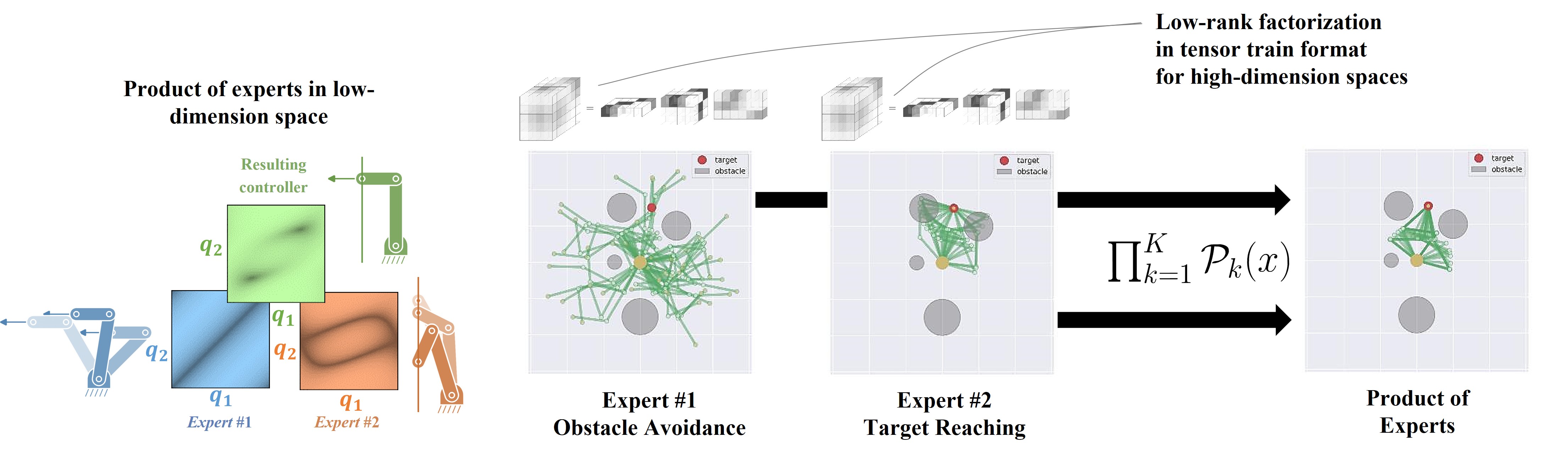}
     \caption{Examples of utilizing multiple experts to extract the feasible set. \emph{Left}: Combining one expert that aligns the end-effector horizontally and another that positions the end-effector on a desired vertical line. The resulting distribution satisfies both constraints. The matrices show the corresponding configuration distribution with the darker values showing the desired configurations. \emph{Right}: Similarly, combining models for obstacle avoidance and target reaching generates configurations (green lines) that meet both goals.}
     \label{fig:TT-PoE-Exps}
\end{figure*}

Fig.~\ref{fig:TT-PoE-Exps} demonstrates two examples of leveraging multiple expert models to address complex tasks. On the left, one expert models the joint angles for achieving a specific orientation, while another focuses on positions along a desired vertical line. The resulting combined distribution respects both constraints and is more sparse compared to the individual models, as evident from the corresponding matrices shown in the figure. On the right, one expert focuses on obstacle avoidance, while another models the task of reaching the target. The combined distribution ensures that the target is reached while avoiding obstacles, effectively balancing both objectives. This approach highlights the power of combining simpler expert models to manage complex scenarios. Furthermore, the modularity of this method allows constraints to be selectively applied. For instance, in the right example, the obstacle avoidance model can be prioritized for intermediate trajectory points, while the target-reaching distribution is only necessary for the endpoint.

Moreover, this concept extends beyond simple AND operations. Indeed, we can also consider other logical behaviors such as OR operations. Without delving into further details, we can express the distribution that satisfies at least one of the constraints \(f_1\) or \(f_2\) as
\begin{equation}
    \bm{\mathcal{P}}^f = \bm{\mathcal{P}}^{f_1} + \bm{\mathcal{P}}^{f_2} - \bm{\mathcal{P}}^{f_1} * \bm{\mathcal{P}}^{f_2},
\end{equation}
These operations can further assist in solving more complex problems by focusing on simpler ones, allowing us to efficiently utilize the learned models instead of relearning them for each situation.

\section{Results}
\label{sec:res}
We applied our methodology to several tasks, verifying the effectiveness of the proposed approach by conducting a comparative analysis with other baselines, namely MPPI, Proj-MPPI (when possible) and NFs-PoE-MPPI. In the MPPI baseline, constraints are integrated directly into the main cost function (i.e., as relaxed constraints), which aligns with the most widely accepted approach in the literature employing this method. In Proj-MPPI, each sample generated by the MPPI method was initially projected into the feasible region before being processed in the MPPI pipeline. The NFs-PoE-MPPI method leverages normalizing flows (NFs) \citep{dinh2016density} to learn the feasibility distribution, taking inspiration from FlowMPPI \citep{power2023variational}. To learn this distribution with NFs, we used the same amount of data as for TT, ensuring that all data points were positive. We maintained consistency in the number of data points across methods to enable a fair comparison. Details regarding the hyperparameters used in various learning methods across different tasks are provided in Table \ref{tab:learning_data}. 

The core problem, particularly the optimality distribution, is identical for all baselines. Specifically, each method employs a high-penalty term with the same value within the cost function to reject infeasible samples. This mechanism is critical for the constrained tasks considered in this paper; without it, even a few infeasible samples that score well on the unconstrained cost could degrade performance. This approach is consistent with prior work~\citep{Jankowski23ICRA, bhardwaj2022storm, power2023variational}. When comparing baselines, we therefore focus on evaluating the unique components of each method, such as the feasibility expert in PoE-MPPI, rather than the shared rejection mechanism. This separation allows us to independently assess the contribution of these additional components.

The comparison is based on three metrics: the success rate, the total cost, and the number of steps required to reach the goal. To ensure fair comparison despite the varying complexities of different task parameters, we normalize the results of each baseline method to those of the MPPI method. We then report the mean for the logarithms of these normalized values. Thus, negative values indicate better performance relative to the MPPI method.

Additionally, we conduct experiments with different sampling regimes, specifically with 16 samples for a low-sampling regime, 64 samples for a medium-sampling regime, and 512 samples for a high-sampling regime. Further information regarding the experiments, including the definition of the cost function and hyperparameter details, is available in the Appendix.

Finally, we demonstrate that our framework scales to high-dimensional problems and can seamlessly integrate with alternative optimality distributions, such as the diffusion-based DIAL-MPC \citep{xue2025icra}.
\subsection{Obstacle Avoidance (PNGRID)}
\begin{figure}[tb!]
    \centering
        \subfigure[]{%
        \includegraphics[height=0.2\textwidth]{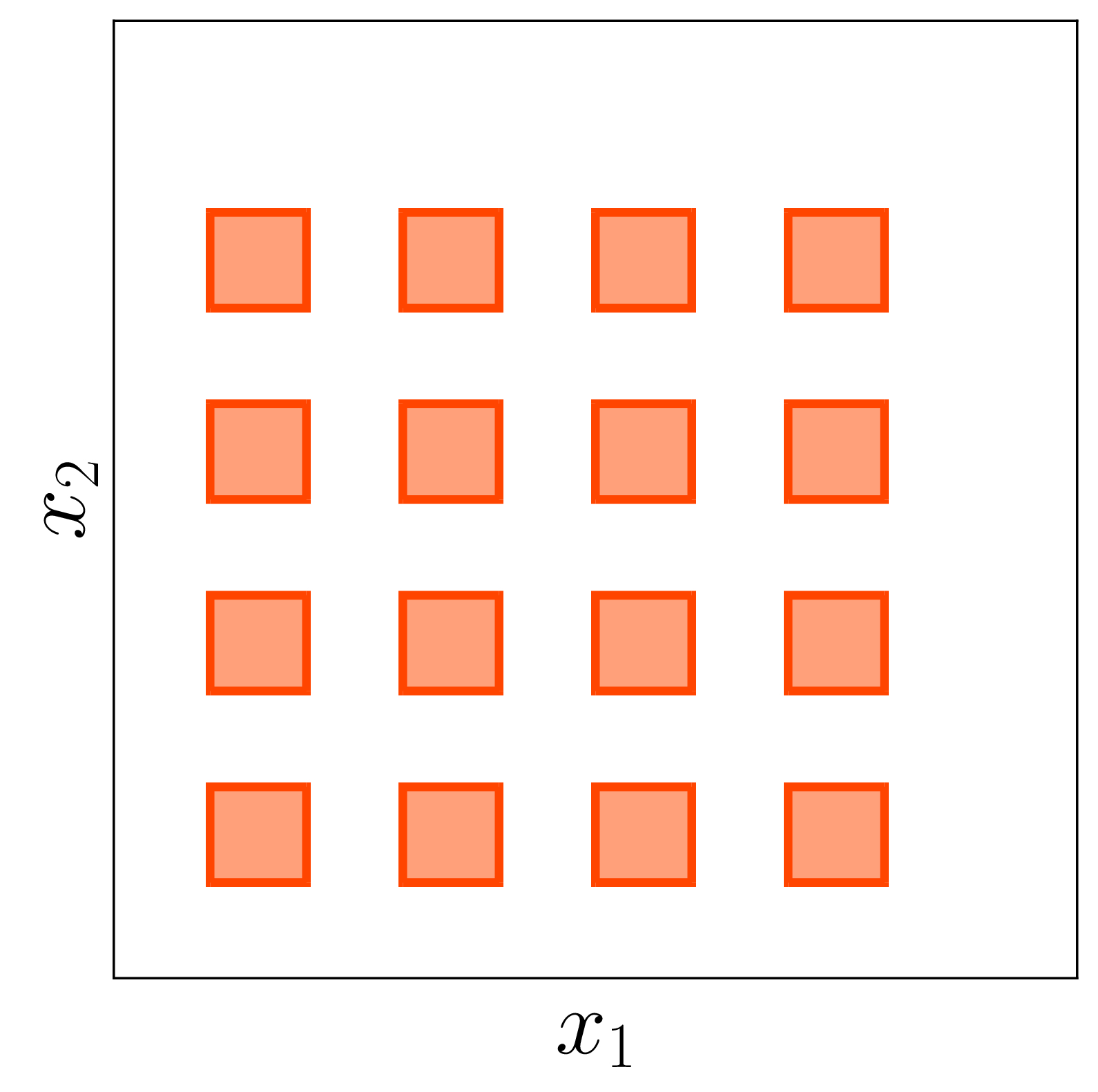}
        }
    \hfill
    \subfigure[]{%
        \includegraphics[height=0.2\textwidth]{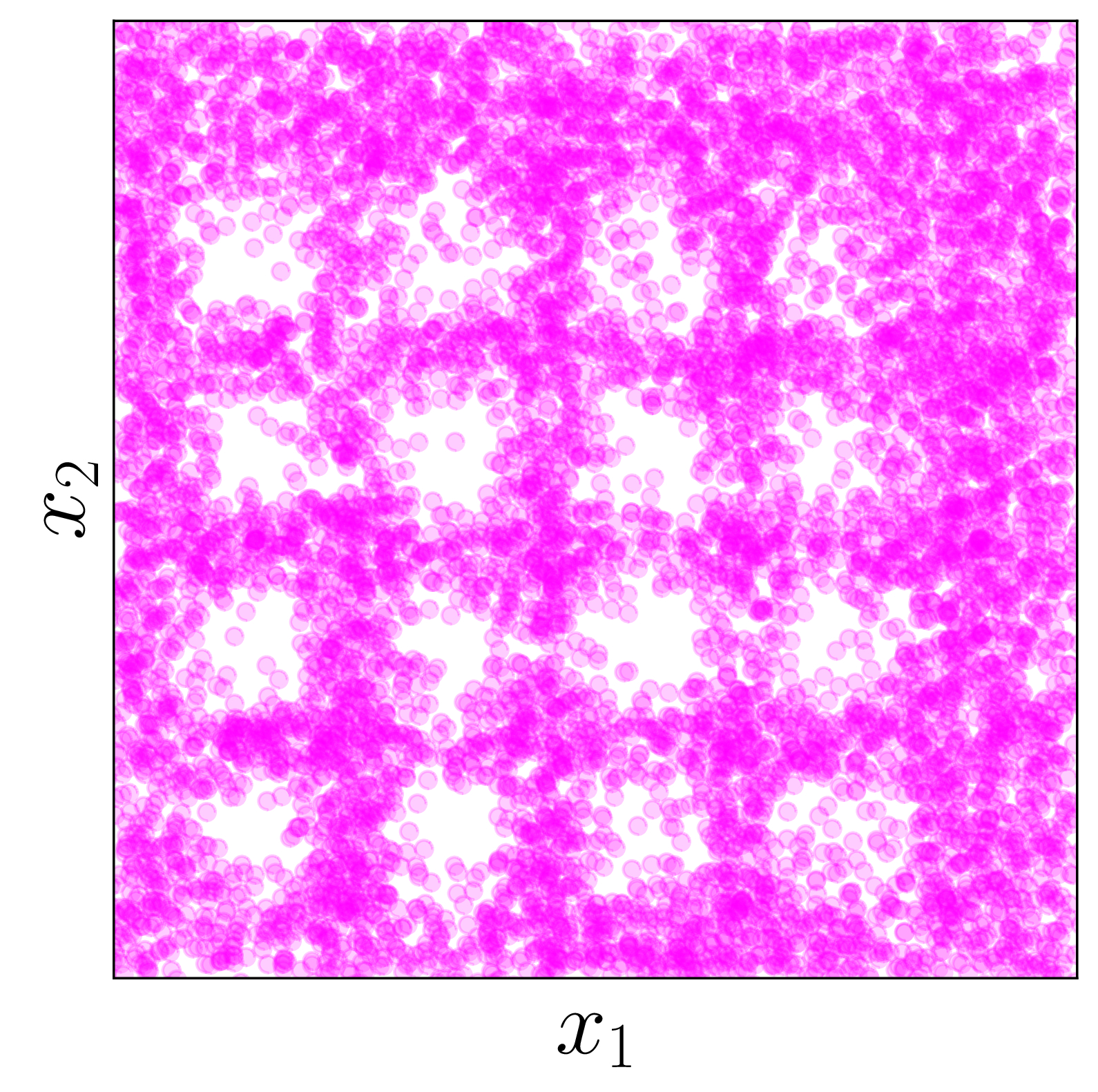}
        }
    \hfill
    \subfigure[]{%
        \includegraphics[height=0.2\textwidth]{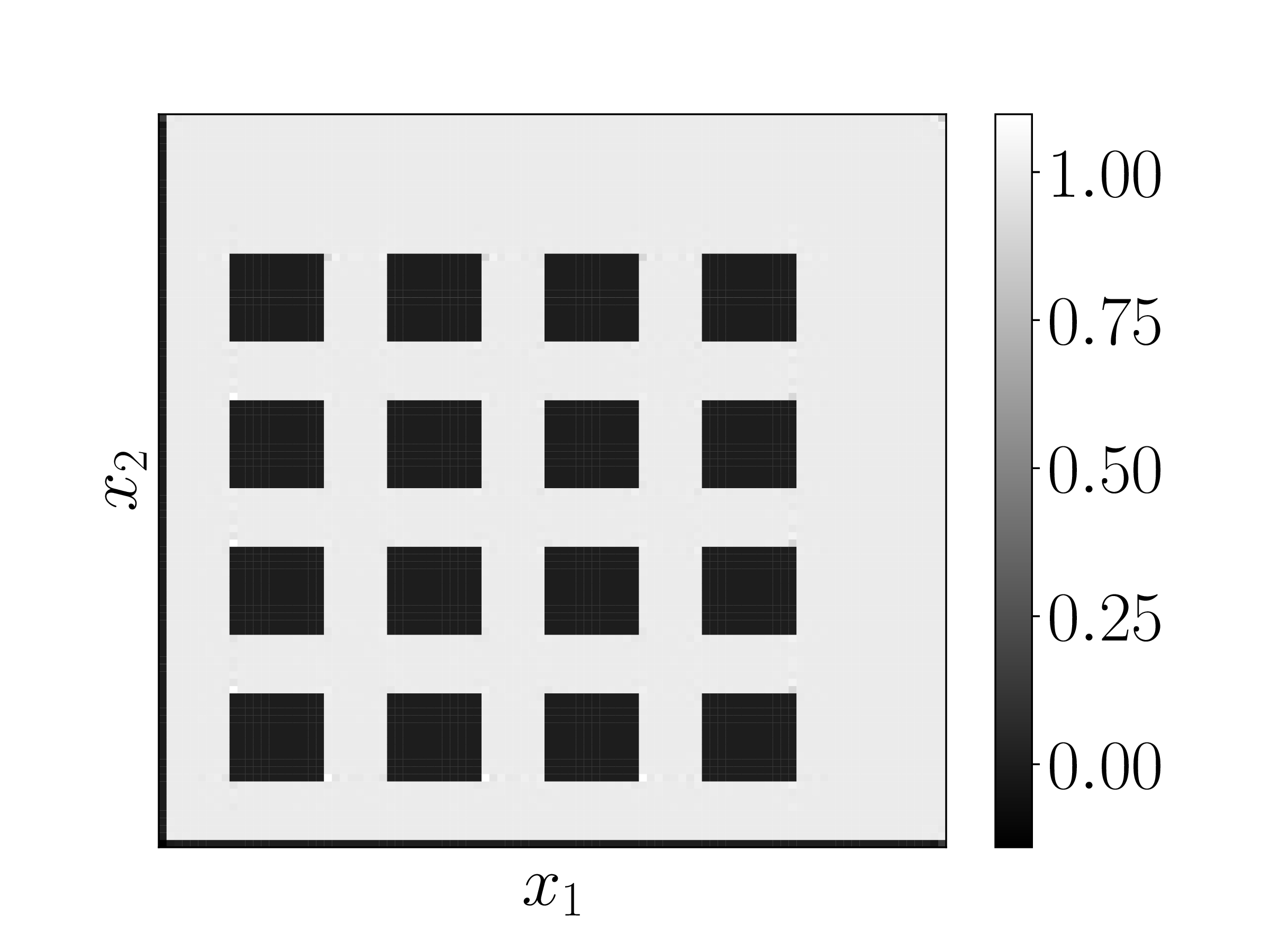}
        }
    \hfill
    \subfigure[]{%
    \includegraphics[height=0.2\textwidth]{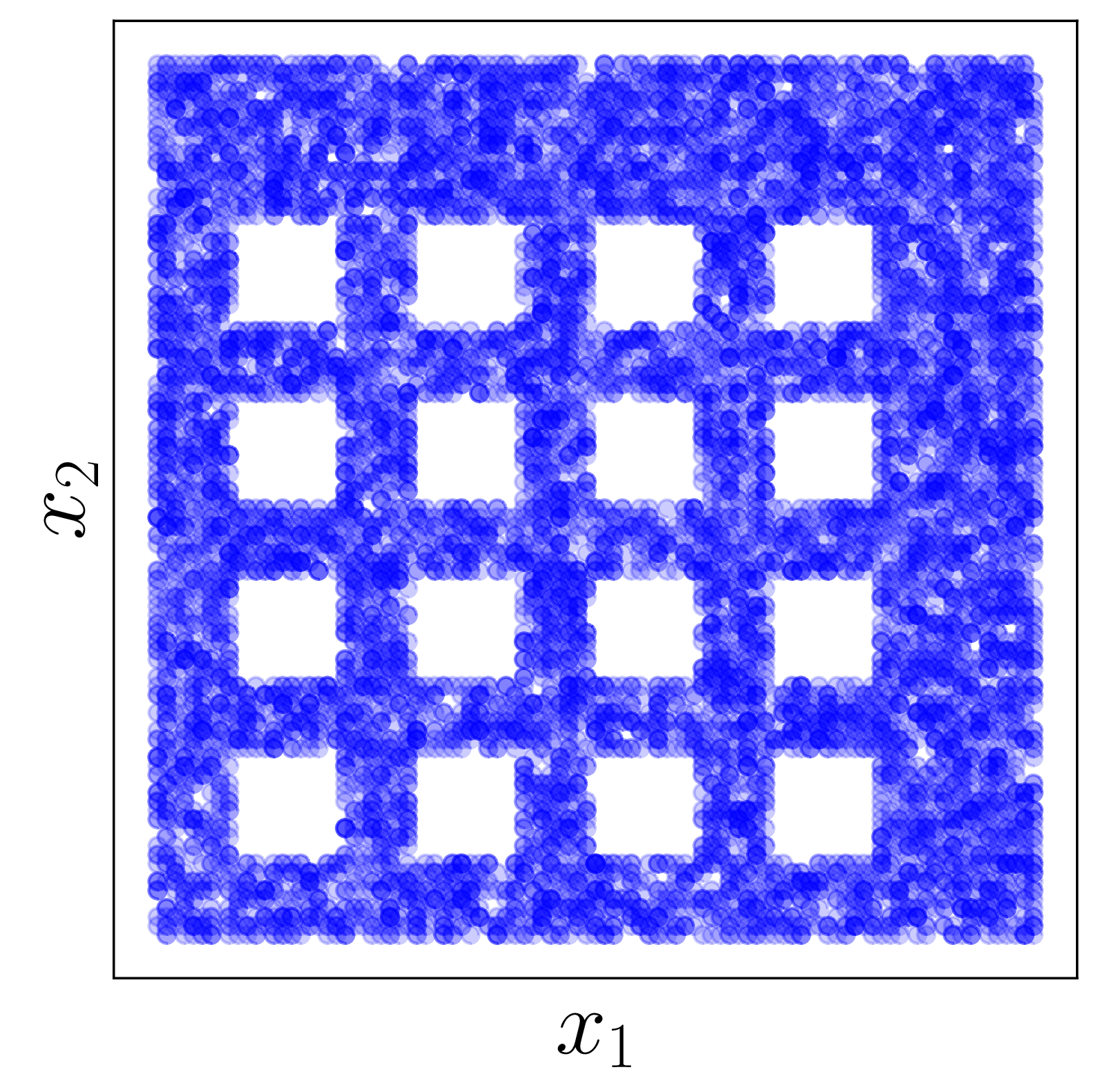}
    }
    \caption{PNGRID Configuration. (a) Depiction of the primary environment, with orange rectangles representing obstacles. (b) Samples generated from a GMM model using $k=400$ Gaussian components, trained with expectation-maximization (EM) algorithm. (c) Distribution learned through the TT-SVD method, and the samples generated from this distribution are depicted in (d). As evident from the results, the TT-SVD method precisely captures the feasible set, with the generated samples exhibiting a higher concentration within the feasible area compared to those generated by the GMM model.}
    \label{fig:pnr_grid_env}
\end{figure}

The PNGRID task, as utilized in prior studies like \citep{sacks2023learning}, is an instance of the widely studied obstacle-avoidance problem in the SMPC field. This task, illustrated in Fig. \ref{fig:pnr_grid_env}-a, involves a 2D environment where an agent must navigate from a randomly selected start point to a randomly assigned target while avoiding static obstacles marked in orange. For this task, the feasibility distribution captures the obstacle-free areas, directing sampling toward viable paths. 

Table \ref{tab:experiment_proj} presents a comparative analysis of system performance across baseline approaches, based on averages from 100 trials with random start and target points. All Proj-MPPI, NFs-PoE-MPPI and TT-PoE-MPPI demonstrate performance improvements over the MPPI method, indicated by negative values, consistent with prior findings \citep{power2023variational, sacks2023learning}. Notably, the TT-PoE-MPPI method consistently outperforms the other methods, particularly in terms of success rate within the medium to low sampling regime. However, as the number of samples increases, both NFs-PoE-MPPI and TT-PoE-MPPI exhibit improved performance in terms of step efficiency and cost values. In contrast, Proj-MPPI's performance aligns more closely with MPPI under high-sampling conditions than in the lower and medium sampling regimes. This trend highlights the competitive advantage of the TT-PoE-MPPI and NFs-PoE-MPPI approaches across varying sampling densities.

\begin{table*}
\centering
\caption{Comparison of various methods across different tasks employing different sampling regimes. Values are normalized to the MPPI method. Hence, the number of steps and cost values for the MPPI method are not provided. Negative values indicate superior performance compared to the MPPI method, with bold formatting indicating superior results.}
\resizebox{\textwidth}{!}{
\begin{tabular}{|c|c|c|c|c|c|c|c|c|c|c|c|} 
\label{tab:experiment_proj}
     \multirow{2}{*}{Task} & \multirow{2}{*}{\#samples} & MPPI & \multicolumn{3}{c|}{Proj-MPPI} &\multicolumn{3}{c|}{NFs-PoE-MPPI} & \multicolumn{3}{c|}{TT-PoE-MPPI (proposed)} \\
     \cline{3-12}
     & &succ. & succ. & log($\#\text{steps}_n$) & log($\text{cost}_n$) & succ. & log($\#\text{steps}_n$) &log($\text{cost}_n$)& succ. & log($\#\text{steps}_n$) &log($\text{cost}_n$) \\
     \Xhline{5\arrayrulewidth}
     \multirow{3}{*}{{\small{PNGRID}}} & 16 & $46\%$ &$70\%$ & $-0.07$ & $-0.19$& $74\%$ & $-0.03$ & $-0.04$ & $\bm{96\%}$ & $\bm{-0.81}$ & $\bm{-0.35}$\\
     \cline{2-12}
      & 64 & $81\%$ & $92\%$ & $-0.06$ & $-0.09$ & $90\%$ & $-0.10$ & $-0.16$ & $\bm{100\%}$ & $\bm{-0.69}$ & $\bm{-0.90 }$\\
    \cline{2-12}
     & 512 & $93\%$ & $96\%$ & $-0.06$ & $-0.09$ & $97\%$ & $-0.13$ & $-0.16$ & $\bm{100\%}$ & $\bm{-0.65}$ & $\bm{-0.78}$\\
     \Xhline{5\arrayrulewidth}
  \multirow{3}{*}{\small Sphere. Man.} & 16 & $73\%$ & $73\%$ & $-0.11$ & $-0.09$& $75\%$ & $0.01$ & $-0.57$ & $\bm{100\%}$ & $\bm{-0.32}$ & $\bm{-1.08}$ \\
     \cline{2-12}
      & 64 & $93\%$ & $77\%$ & $0.03$ & $-1.57$ &  $87\%$ & $0.20$ & $-0.68$ & $\bm{100\%}$ & $
      \bm{-0.39}$ & $\bm{-4.38}$\\
    \cline{2-12}
     & 512 & $\bm{100\%}$ & $78\%$ & $0.05$ & $-1.89$ & $\bm{100\%}$ & $-0.26$ & $-1.97$ & $\bm{100\%}$ & $
      \bm{-0.89}$ & $\bm{-3.08}$\\
    \Xhline{5\arrayrulewidth}
  \multirow{3}{*}{\small Sin. Man.} & 16 & $31\%$ & $38\%$  & $-0.09$  & $-1.08$ & $38\%$  & $-0.15$  & $-0.08$ & $\bm{100\%}$ & $\bm{-1.58}$ & $\bm{-1.86}$\\
     \cline{2-12}
      & 64 & $44\%$ & $44\%$  & $-0.09$ & $-0.87$ & $50\%$  & $-0.41$ & $-0.40$ & $\bm{100\%}$  & $\bm{-2.37}$& $
      \bm{-2.03}$ \\
    \cline{2-12}
     & 512 & $55\%$ & $68\%$  & $-0.30$ & $-0.68$& $65\%$  & $-0.52$ & $-0.14$ & $\bm{100\%}$ & $\bm{-1.92}$ & $
      \bm{-2.40}$ \\
    \Xhline{5\arrayrulewidth}
\end{tabular}}
\end{table*}
\subsection{Tracking Tube (e.g., surface cleaning or painting)} \label{sec:manifold}
\begin{figure}[tb!]
   \centering
    \subfigure[]{
        \includegraphics[height=0.23\textwidth]{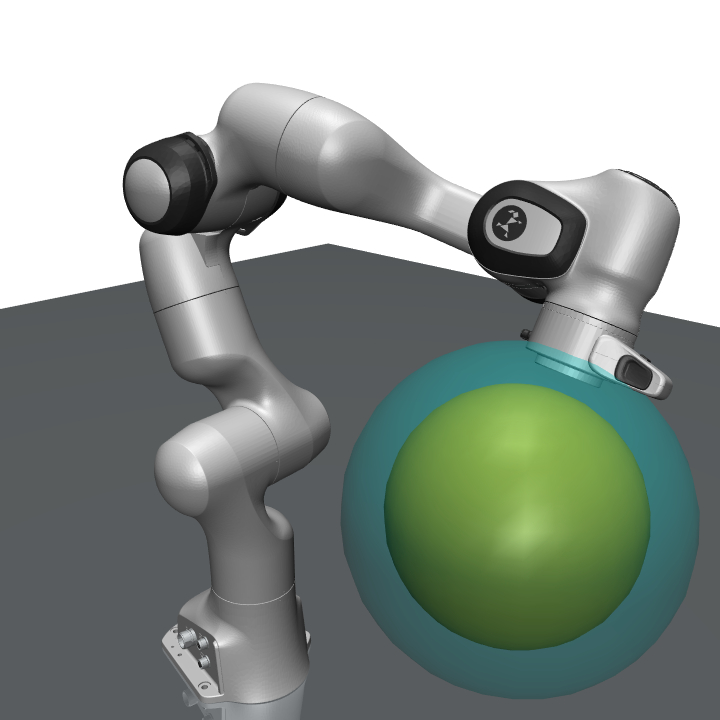}
        }
    \hfill
    \subfigure[]{
        \includegraphics[height=0.23\textwidth]{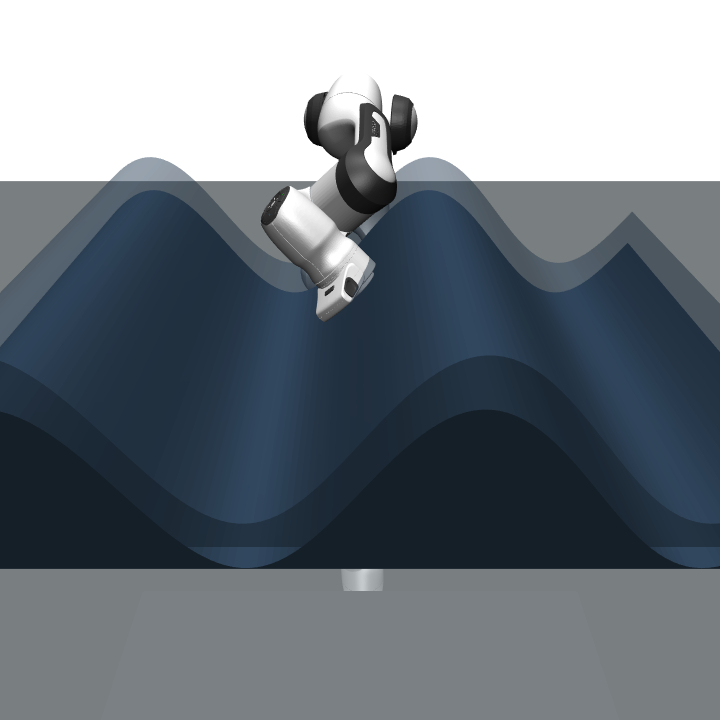}}
        \hfill
    \subfigure[]{
        \includegraphics[height=0.2\textwidth]{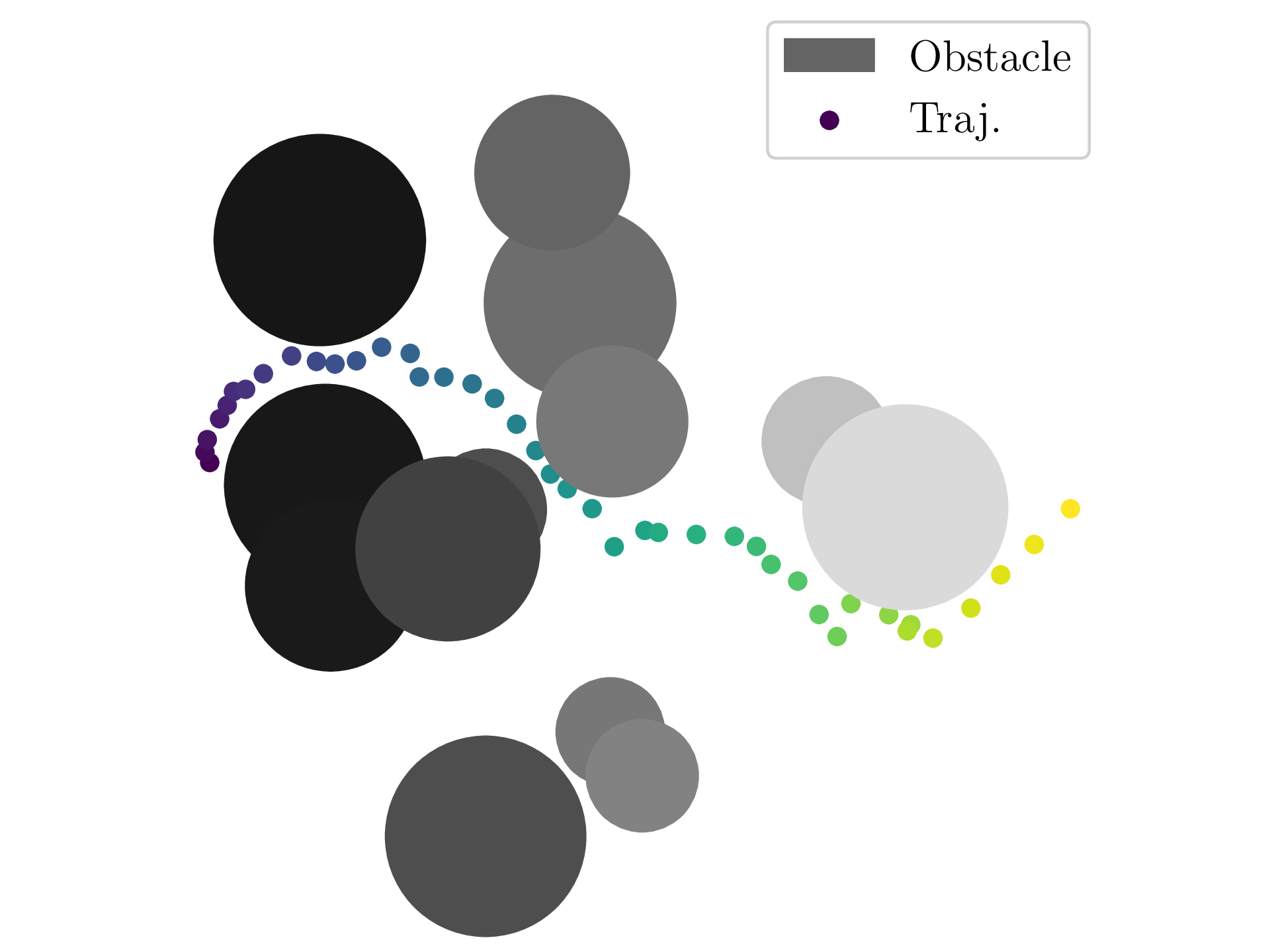}
        }
     \caption{Different environmental setups: (a) sphere tube tracking, (b) sinusoidal wave tube tracking, and (c) Online obstacle-avoidance. In (c), the obstacles become more apparent as the robot approaches them; lighter colors indicate obstacles that appear later in the task.}
     \label{fig:manifold}
\end{figure}

The proposed method is also applied to a task where the robot’s end-effector must traverse a restricted volume (e.g., a spherical or a sinusoidal manifold with some margins, as illustrated in Fig. \ref{fig:manifold}). Such tasks are prevalent in robotics. For instance, when a robot needs to clean or paint a surface or follow a specific path. In this setup, we control the robot's end-effector position in the task space, ensuring it stays within a designated shaded area. However, the feasible region for this task is substantially restricted relative to the entire workspace, making it challenging to sample feasible actions. We employed two different volumes: one between two spheres and another between two sinusoidal waves (Fig. \ref{fig:manifold}). The evaluation results, shown in Table \ref{tab:experiment_proj}, reveal that the proposed method significantly improves the task success rate, especially in low-sampling scenarios. Moreover, our method achieves faster convergence with lower associated costs. As shown in the table, Proj-MPPI exhibits behavior similar to MPPI in terms of success rate (and even lower in the sphere case) and number of steps, while achieving significantly lower cost values. This outcome is likely due to the projection step, which ensures that nearly all samples remain within the designated area, leading to reduced costs. However, this projection mechanism appears to make the system more inclined to stay within its current location rather than actively exploring new solutions, which may contribute to its lower success rate on the sphere manifold compared to MPPI. In contrast, NFs-PoE-MPPI shows markedly lower performance than TT-PoE-MPPI, for reasons that are discussed in detail in Section \ref{sec:discuss}.

In this experiment, our objective was to evaluate the performance of our systems on equality constraints, which in this case correspond to a manifold. However, we acknowledge that the TT method, which is a volumetric representation, cannot directly enforce strict equality constraints. To address this limitation, we evaluated the performance of our approach on a relaxed version of the problem by introducing margin boundaries around the manifold, effectively creating a "soft" constraint region for analysis. We then evaluate how the pipeline behaves as this margin is progressively reduced, and benchmark it against an intrinsic‐sampling approach (i.e., Riemannian log/exp maps), which guarantees constraint satisfaction intrinsically by sampling in the tangent space first, then mapping to the task space. For certain manifolds, like a sphere, these mappings can be computed analytically (see Appendix for details). 

Table~\ref{tab:varied_margin} presents the results of this comparison. We evaluate our proposed method against the Riemannian log/exp maps approach (specifically for a spherical manifold) and a vanilla-MPPI method. As shown in Table~\ref{tab:varied_margin}, when the margin exceeds the TT discretization (2 cm), our TT‑PoE‑MPPI method outperforms both standard MPPI and the Riemannian mapping approach. We attribute this to the fact that intrinsic sampling in the tangent space incurs inefficiencies, particularly when shifting between time steps and averaging, whereas the TT representation captures the volumetric structure more globally. As the margin shrinks, however, the intrinsic method ultimately yields the best performance, although our method remains superior to vanilla MPPI.

These results suggest that TT‑PoE‑MPPI can accelerate the dynamic sampler under equality constraints, but that pure intrinsic sampling may be preferable for very tight margins. Moreover, TT‑PoE can be combined with intrinsic approaches: once an intrinsic transform renders the problem unconstrained, additional constraints (e.g., joint limits or other inequalities) can still be enforced by defining the TT experts in the transformed space. Note tha the optimality distributions for the intrinsic sampler and for other methods inhabit different spaces in this experiment. Hence, as the margin shrinks to zero, the unconstrained intrinsic sampler can serve as the ground-truth solution to the constrained problem. Furthermore, our product‑of‑experts framework consistently improves sampling efficiency without imposing additional structural assumptions, such as that the constraint space forms a Riemannian manifold or that the constraints are affine in the action variables, as required by \citet{CariusEtAl2021}.

\begin{table*}
\centering
\caption{Comparison of different methods 
 on sphere tube-tracking task across varying margin sizes. The intrinsic MPPI method (MPPI solved directly on the spherical manifold using log/exp maps) reformulates the problem as unconstrained and, in the limit as the margin~$\to0$, serves as the \emph{ground truth} for the other baselines. Values are normalized to the MPPI method. Hence, the number of steps and cost values for the MPPI method are not provided. Negative values indicate superior performance compared to the MPPI method, with bold formatting indicating superior results.}
{\color{black}
\begin{tabular}{|c|c|c|c|c|c|c|c|} 
\label{tab:varied_margin}
     \multirow{2}{*}{Margin size} &  MPPI & \multicolumn{3}{c|}{TT-PoE-MPPI} &\multicolumn{3}{c|}{Intrinsic MPPI}\\
     \cline{2-8}
     & succ. rate& succ. rate& log($\#\text{steps}_n$) & log($\text{cost}_n$) & succ. rate & log($\#\text{steps}_n$) &log($\text{cost}_n$)\\
     \Xhline{5\arrayrulewidth}
    5 [cm] & ${98\%}$ & $\bm{100\%}$& $\bm{-0.98}$ & $\bm{-2.94}$ & $\bm{100\%}$ & $-0.72$ & $-2.55$\\
    \Xhline{5\arrayrulewidth}
    3 [cm] & ${91\%}$ & $\bm{100\%}$& $\bm{-1.06}$ & $\bm{-4.03}$ & $\bm{100\%}$ & $-0.81$ & $-3.78$\\
    \Xhline{5\arrayrulewidth}
    1 [cm] & ${78\%}$ & $92\%$& $-0.02$ & $-4.91$ & $\bm{100\%}$ & $\bm{-0.37}$ & $\bm{-5.73}$\\
    \Xhline{5\arrayrulewidth}
\end{tabular}}
\end{table*}

\subsection{Non-prehensile Manipulation (Planar-pushing)}
An other task that we consider is a pushing task \citep{mason1986mechanics, Xue23ICRA}, a type of non-prehensile, contact-rich manipulation. These tasks are inherently challenging as they require the robot to handle a range of contact models, with hybrid dynamics that complicate the use of gradient-based methods. This hybrid nature arises from various factors, including the geometry of objects, the contact modes, as well as the task level. In this work, we focus specifically on the challenges related to geometry and contact mode, while task-level hybridization has been more extensively explored within task and motion planning (TAMP) frameworks \citep{Hollady23IJRR}. 

Previous studies have attempted to address the hybrid nature of contact through unified contact modes \citep{anitescu2006optimization, manchester2020variational, posa2014direct,yunt2006SUMT} or by employing mixed-integer programming (MIP) to jointly optimize hybrid and continuous variables \citep{aceituno2022hierarchical}. However, these approaches struggle with geometric hybridness, and solving complex non-linear, non-convex problems with MIP can be computationally intensive. SMPC offers an effective alternative \citep{pezzato2023sampling}, as it avoids reliance on gradients. Furthermore, as demonstrated by \cite{pang2023global}, the inherent ability of sampling-based methods to smooth functions proves advantageous in these hybrid and contact-rich scenarios.

Using the method developed in this work, the optimality expert can focus on pushing an object toward a goal while adapting to environmental changes, such as the introduction of obstacles that were not present in the training data. The feasibility distribution is designed to model the efficiency of the actions taken. As mentioned earlier, certain actions are inconsequential if they lead the robot away from the object or otherwise fail to move the object meaningfully. In an offline phase, we gather data on various state-action combinations, labeling them as efficient if the robot moves toward the object or if the action causes the object to relocate. Specifically, \( p_{\text{feas}}(\bm{x}_t, \bm{u}_t) \) in \eqref{eq:const_static} returns 1 if the action \( \bm{u}_t \) in state \( \bm{x}_t \) is effective and 0 otherwise. This approach is similar to embedding a cost function that encourages the robot to approach the object. However, we demonstrate that our method is more efficient than relying on a penalty term within the cost function.
 
\begin{figure}[tb!]
    \centering
    \subfigure[]{
        \includegraphics[height=0.23\textwidth]{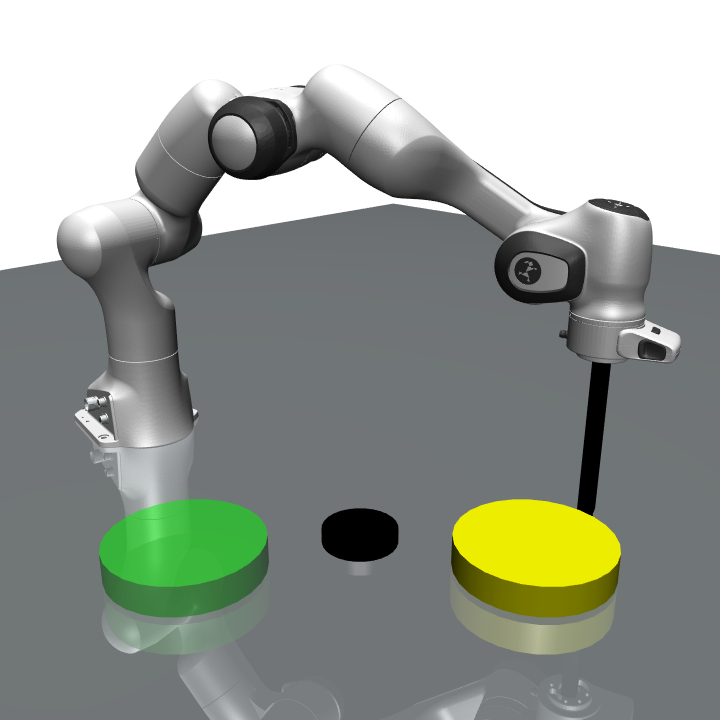}
        }
    \hfill
    \subfigure[]{
        \includegraphics[height=0.23\textwidth]{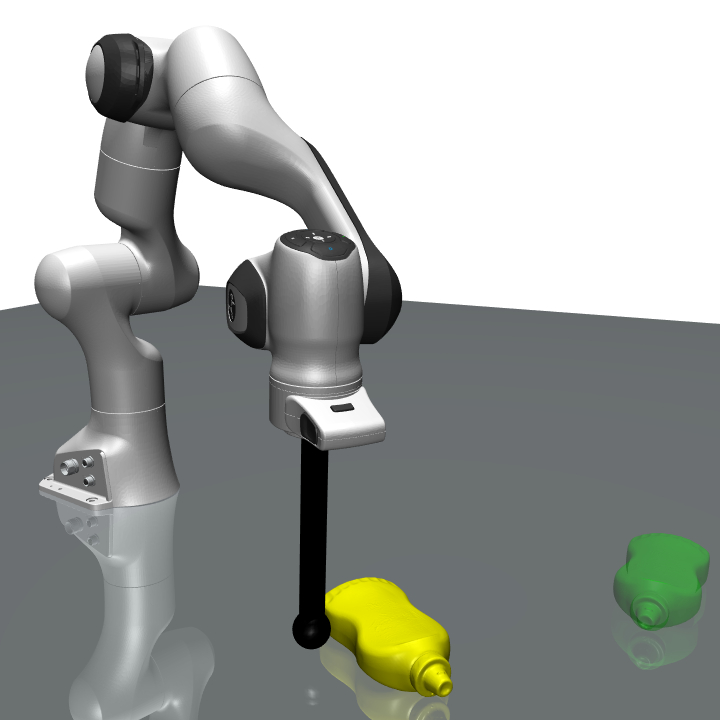}
        }
    \hfill
    \caption{Different environment setups for pushing various objects: (a) a cylinder and (b) a mustard bottle. The obstacle is shown in black, while the green objects represent the target locations.}
    \label{fig:env_vis}
\end{figure}

In this experimental setup, we employed three objects: a cylinder (Fig.~\ref{fig:env_vis}-a), a box, and a mustard bottle sourced from the YCB dataset \citep{calli2017yale} (Fig.~\ref{fig:env_vis}-b). Additionally, for the cylinder scenario, we introduced an obstacle into the environment to assess the efficacy of the methods in navigating around obstacles. It is important to note that the feasibility distribution is unaware of the obstacle in the environment (not considered in the training), as addressing this situation falls within the responsibility of the optimality distribution. 

The results for these experiments are shown in Table \ref{tab:experiment}. As can be seen, our approach exhibits enhanced performance when compared to both NFs-PoE-MPPI and MPPI methodologies, demonstrated through superior cost values and a reduced number of steps. Moreover, the proposed method attains a superior success rate compared to other baseline methods. 

\begin{table*}
\centering
\caption{Comparison of various methods across different tasks employing different sampling regimes. Values are normalized to the MPPI method; hence, the number of steps and cost values for the MPPI method are not provided. Negative values indicate superior performance compared to the MPPI method, with bold formatting indicating superior results.}
\resizebox{\textwidth}{!}{
\begin{tabular}{|c|c|c|c|c|c|c|c|c|} \label{tab:experiment}
     \multirow{2}{*}{Task} & \multirow{2}{*}{\#samples} & MPPI & \multicolumn{3}{c|}{NFs-PoE-MPPI} & \multicolumn{3}{c|}{TT-PoE-MPPI} \\
     \cline{3-9}
     & &succ. rate & succ. rate & log($\#\text{steps}_n$) & log($\text{cost}_n$) & succ. rate & log($\#\text{steps}_n$) &log($\text{cost}_n$) \\
     \Xhline{5\arrayrulewidth}
     \multirow{3}{*}{Push. Cir.} & 16 & $89\%$ & $\bm {96\%}$ & $-0.19$ & $-0.22$ & $95\%$ & $\bm{-0.35}$ & $\bm{-0.41}$\\
     \cline{2-9}
      & 64 & $96\%$ & $99\%$ & $-0.27$ & $-0.30$ & $\bm{100\%}$ & $\bm{-0.38}$ & $\bm{-0.45}$\\
    \cline{2-9}
     & 512 & $97\%$ & $\bm{100\%}$ & $-0.24$ & $-0.27$ & $\bm{100\%}$ & $\bm{-0.38}$ & $\bm{-0.45}$\\
     \Xhline{5\arrayrulewidth}
     \multirow{3}{*}{Push. Box} & 16 & $\bm{100\%}$ & $\bm{100\%}$ & $0.29$ & $0.28$ & $\bm{100\%}$ & $\bm{-0.17}$ & $\bm{-0.21}$ \\
     \cline{2-9}
      & 64 & $\bm{100\%}$ & $\bm{100\%}$ & $-0.06$ & $-0.08$ & $\bm{100\%}$ & $
      \bm{-0.25}$ & $\bm{-0.29}$\\
    \cline{2-9}
     & 512 & $\bm{100\%}$ & $\bm{100\%}$ & $0.25$ & $0.23$ & $\bm{100\%}$ & $\bm{-0.01}$ & $\bm{-0.03}$\\
     \Xhline{5\arrayrulewidth}
     \multirow{3}{*}{Push. Mus.} & 16 & $70\%$ & $73\%$ & $0.12$ & $0.11$ & $\bm{89\%}$ & $\bm{-0.21}$ & $\bm{-0.25}$\\
     \cline{2-9}
      & 64 & $75\%$ & $90\%$ & $-0.01$ & $-0.02$ & $\bm{91\%}$ & $
      \bm{-0.33}$ & $\bm{-0.34}$\\
    \cline{2-9}
     & 512 & $75\%$ & $90\%$ & $0.01$ & $-0.01$ & $\bm{91\%}$ & $\bm{-0.26}$ & $\bm{-0.28}$\\
     \Xhline{5\arrayrulewidth}
\end{tabular}}
\end{table*}

\begin{figure*}[bt]
    \centering
    \subfigure[]{
        \includegraphics[height=0.23\textwidth]{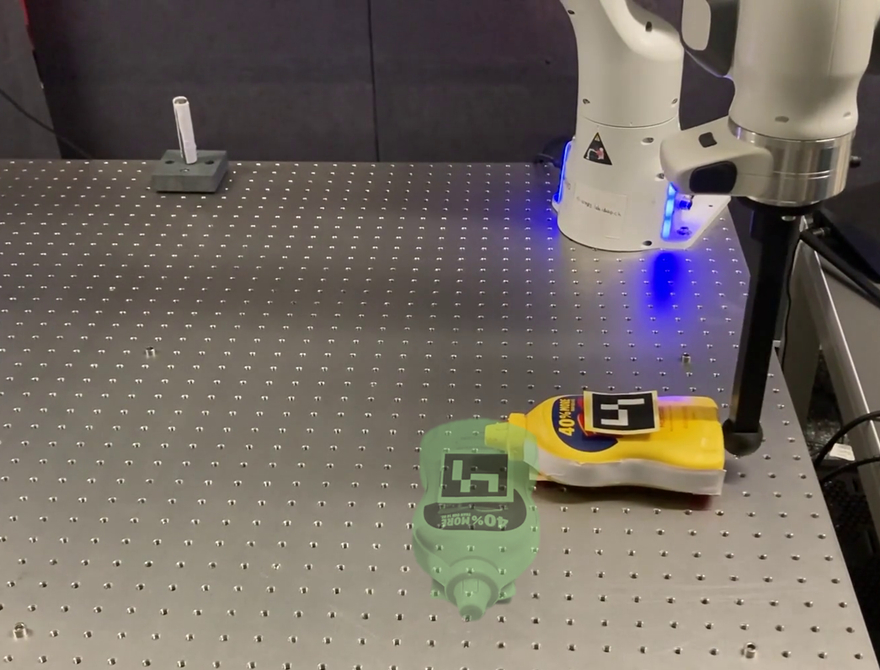}
        }
    \hfill
    \subfigure[]{
        \includegraphics[height=0.23\textwidth]{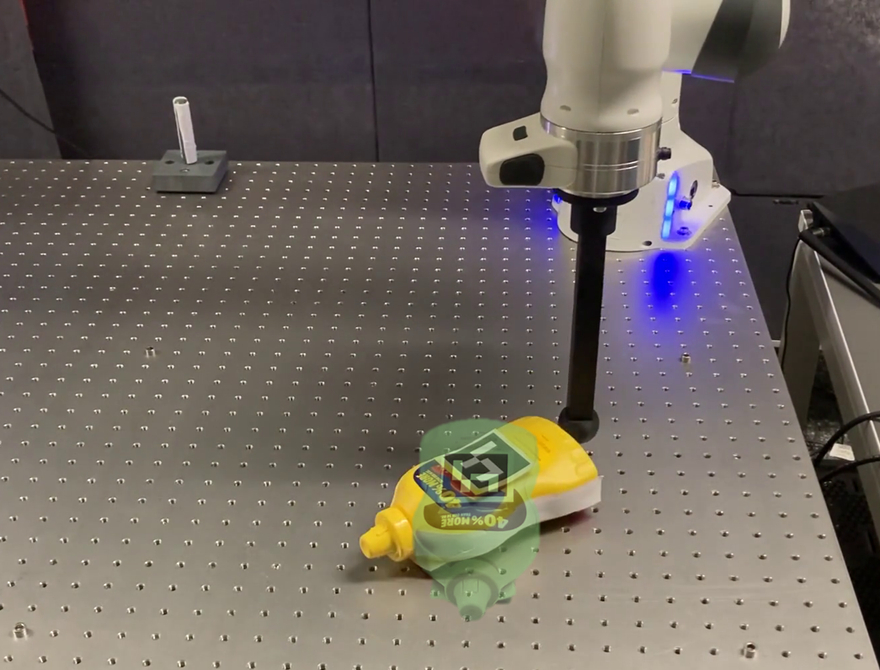}
        }
            \hfill
    \subfigure[]{
        \includegraphics[height=0.23\textwidth]{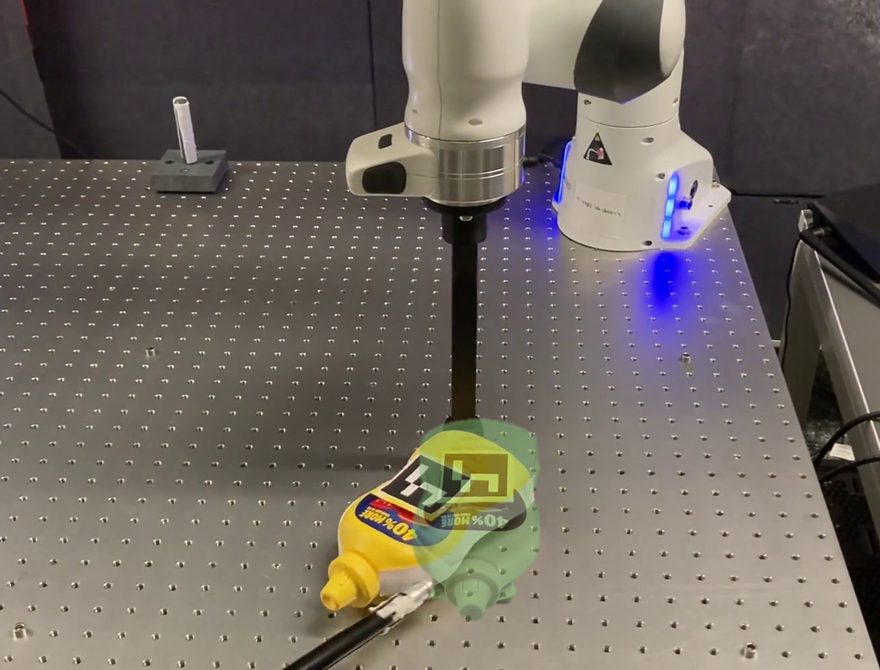}
        }
            \hfill
    \subfigure[]{
        \includegraphics[height=0.23\textwidth]{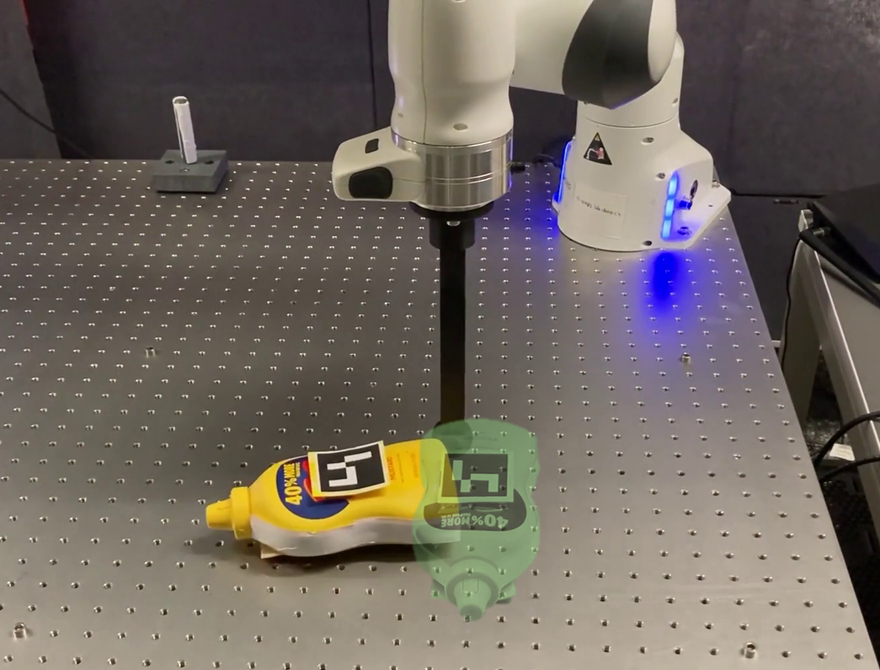}
        }
            \hfill
    \subfigure[]{
        \includegraphics[height=0.23\textwidth]{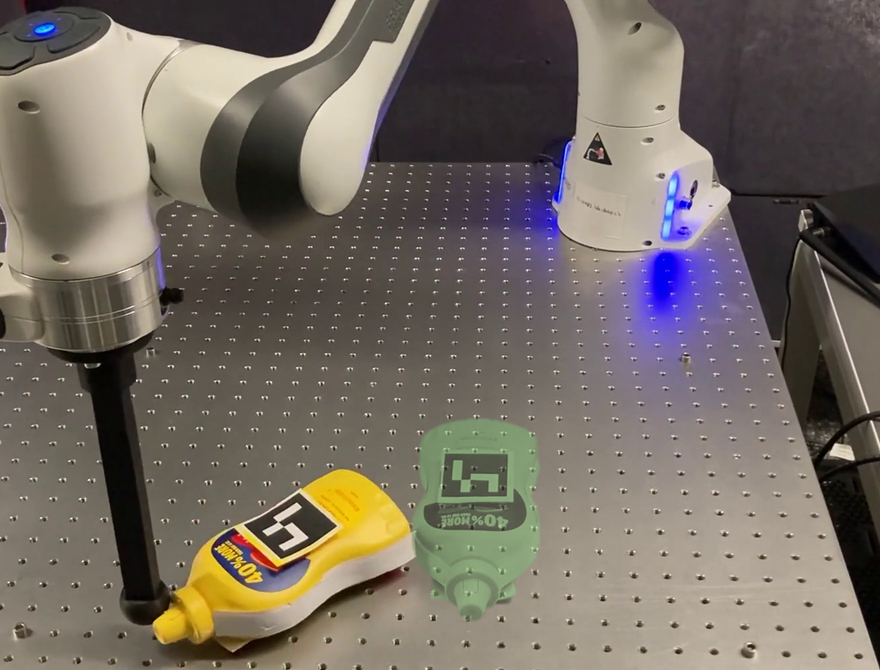}
        }
                    \hfill
    \subfigure[]{
        \includegraphics[height=0.23\textwidth]{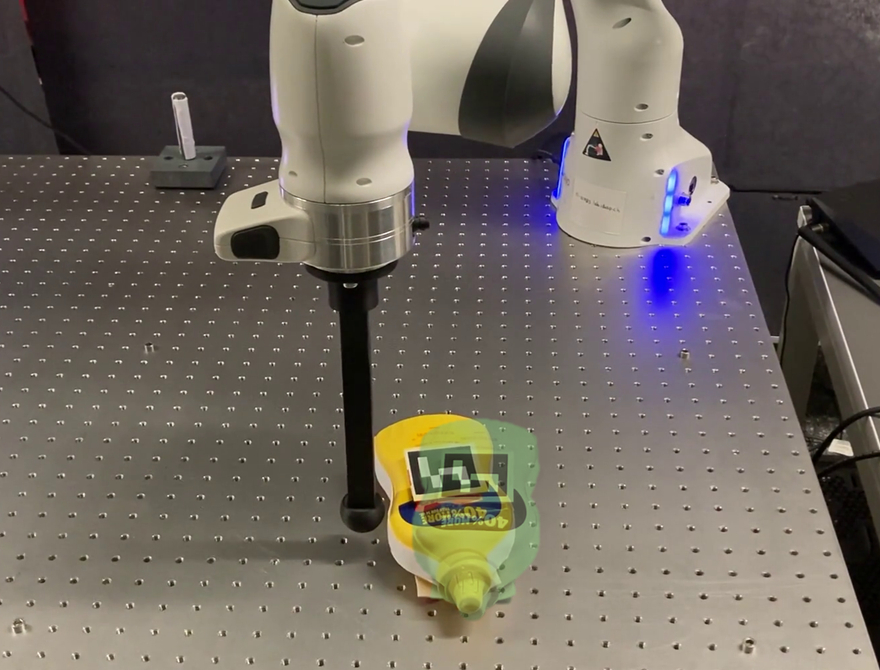}
        }
    \caption{Frames selected from the real robot experiment showing the process of pushing a mustard bottle towards the target (shown as the semi-transparent picture). An external disturbance was introduced to the system at frame (c). However, the planner effectively navigates this scenario, successfully guiding the object toward the target.}
    \label{fig:real_robot}
\end{figure*}

The proposed method was further applied to the real robot for pushing a mustard bottle using only 32 samples which is very low compared to the commonly used number of samples in other SMPC methods. We chose this low number deliberately to show the capability of our method in increasing the efficiency of samples. Selected frames from this experiment are illustrated in Fig. \ref{fig:real_robot}. The rationale behind selecting the mustard bottle lies in its complex shape compared to other objects, thereby serving as a demonstrative example that the shapes of the objects involved do not constrain our method. The objective in this scenario is to guide the object back to the goal location (transparent shape in Fig. \ref{fig:real_robot}), while the object undergoes a 90-degree rotation. The receding horizon controller employed allows the system to effectively handle external disturbances, as demonstrated in the accompanying video. Disturbances were introduced to the system between Fig. \ref{fig:real_robot}-c and \ref{fig:real_robot}-d. Notably, the planner adeptly manages this situation, guiding the object towards the target.

\begin{table}
\centering
\caption{Comparison of TT-PoE-MPPI with MPPI in an online obstacle avoidance task across 100 random obstacle locations. Values are normalized to the MPPI method. Thus, steps and cost values for MPPI are omitted. Negative values indicate superior performance, with bold formatting highlighting the best results.}
\resizebox{\columnwidth}{!}{
\begin{tabular}{c|c|c|c|c} \label{tab:experiment_online}
      \multirow{2}{*}{\#samples} & MPPI & \multicolumn{3}{c}{TT-PoE-MPPI} \\
     \cline{2-5}
     & succ. rate & succ. rate & log(\# $\text{steps}_n$) & log($\text{cost}_n$) \\
     \Xhline{5\arrayrulewidth}
     16 & $73 \%$ & $\bm{84 \%}$ & $\bm{-0.09}$ & $\bm{-0.13}$\\
     64 & $73 \%$ & $\bm{86 \%}$ & $\bm{-0.31}$ & $\bm{-0.44}$\\
     512 & $69 \%$ & $\bm{89 \%}$ & $\bm{-0.46}$ & $\bm{-0.66}$\\
\end{tabular}}
\end{table}

\subsection{Online Obstacle Avoidance}
One significant advantage of using TT-SVD to learn the TT distribution is its capability to process the entire dataset in milliseconds if all data points are available. This is especially beneficial when the robot leverages point cloud data online to identify free space in the environment. This rapid processing allows the system to adapt its distribution quickly if new environmental changes are detected, enabling it to sample updated, obstacle-free trajectories efficiently. To evaluate this capability, we designed a task in which a point-mass robot moves from the left to the right side of the environment. As it progresses, new obstacles appear, shown as lighter regions in Fig. \ref{fig:manifold}-c. Table \ref{tab:experiment_online} presents the results of this experiment, demonstrating that the proposed method consistently outperforms the MPPI method across all measured criteria. Notably, the performance gap between the two methods widens as the number of samples increases.

\subsection{Higher Dimensions and Alternative Optimality Distributions}

Previous experiments addressed low-dimensional problems using TT‑SVD and benchmarked against vanilla MPPI. Here, we demonstrate that our method scales to higher-dimensional settings and can be combined with alternative optimality distributions. We evaluate our method on a task in which a Franka Emika robot must traverse an environment with whole‑body obstacle avoidance, where obstacle locations vary across scenes (as illustrated Fig.~\ref{alg:process}-b).  

Since TT‑SVD becomes impractical for the large tensors required, we instead employ TT‑cross with a fixed maximum rank \(r_{\max}=20\).  This low‑rank approximation reconstructs the full tensor from a subset of entries.  Our approach can be applied in two ways:

\textbf{1. Augmented State Conditioning:}  
    Augment the state with all obstacle positions and, at inference time, condition on the current robot state \(\bm{x}_t\) and obstacle locations \(\bm{x}^o_{1:N_o}\).  We define the joint feasibility distribution as  
    \begin{multline}
      \label{eq:rdf_prob}
      p^f\bigl(\bm{x}^o_{1:N_o}, \bm{x}_t, \bm{u}_t\bigr)
      = \bigl(1 - \exp(-\mathrm{dist}(\bm{x}_t)/0.5)\bigr) \\
        \bigl(1 - \exp(-\mathrm{dist}(\bm{x}_{t+1})/0.5)\bigr),
    \end{multline}
    where \(\mathrm{dist}(\cdot)\) is the minimum distance between the robot body and obstacles computed via the Robot Distance Field (RDF) in \citep{Li24ICRA}.  By using a smooth, distance‑based likelihood rather than a binary collision indicator, we favor configurations that keep the robot at a safe distance and produce a lower‑rank function amenable to TT‑cross learning.

\textbf{2. Additive Feasibility Mixture:}  
    Learn a separate feasibility tensor \(\mathcal{P}^{f_i}\) for only one obstacle, then combine them as
    \[
      \mathcal{P}^f \;=\; \sum_{i=1}^{N_o}\,\mathcal{P}^{f_i}.
    \]

The additive mixture remains smoother with a lower rank representation, which is more amenable to the TT-cross method.  We denote this variant as \emph{TT‑PoE‑MPPI‑prod}.

We further evaluate our framework against DIAL‑MPC \citep{xue2025icra}, an annealed MPPI variant that adapts its covariance using diffusion‑inspired updates \citep{ho2020denoising,chi2024diffusionpolicy}. Although DIAL-MPC is significantly more efficient than vanilla MPPI, it still relies on rejection sampling to eliminate infeasible trajectories—because its experts are fused at the function level and have no direct awareness of the feasible solution space. By contrast, our solution-level fusion lets us steer the sampler toward valid regions before any draws are discarded. To highlight this advantage, we embed DIAL-MPC within our TT-PoE pipeline (denoted \emph{TT-PoE-DIAL}) and show that this hybrid consistently outperforms both standalone DIAL-MPC and our original TT-PoE-MPPI in terms of sampling efficiency and overall task success.

Table~\ref{tab:experiment_tt_cross} reports success rate, total cost, and steps-to-goal for each method.  Integrating TT with both MPPI and DIAL‑MPC increases success rates while reducing cost and required steps.  Notably, the TT‑PoE‑MPPI‑prod variant matches or exceeds the performance of TT‑PoE‑MPPI, validating the effectiveness of learning and summing individual feasibility distributions within a product‑of‑experts framework.  We define task success as reaching the target in fewer than 25 steps with total cost below \(10^6\).  Further experimental details are provided in the Appendix.  

\begin{table}
\centering
\caption{\color{black}Comparison of different methodologies in whole body obstacle avoidance task across 100 random obstacle locations and initial and target configurations using 1024 sample size. Values are normalized to the MPPI method. Thus, steps and cost values for MPPI are zero. Negative values indicate superior performance, with bold formatting highlighting the best results.}
\resizebox{\columnwidth}{!}{\color{black}
\begin{tabular}{|c|c|c|c|} \label{tab:experiment_tt_cross}
      Method & succ. rate & log(\# $\text{steps}_n$) & log($\text{cost}_n$)\\
     \hline
     MPPI & $30\%$ & $0.00$ & $0.00$\\
     
     TT-PoE-MPPI & $53\%$ & $-0.10$ & $-0.09$\\
     
     TT-PoE-MPPI-prod & $50\%$ & $-0.14$ & $-0.11$\\
     
     DIAL & $88\%$ & $-0.35$ & $-0.36$\\
     
     TT-PoE-DIAL & $\bm{89\%}$ & $\bm{-0.50}$ & $\bm{-0.53}$\\
     \hline
\end{tabular}}
\end{table}

\section{Discussion}
\label{sec:discuss}
The primary objective of this article is to enhance the sampling distribution employed in SMPC through the application of products of experts in the solution space. As anticipated, all methods with TT-PoE pipeline (NFs-PoE-MPPI, TT-PoE-MPPI, and TT-PoE-DIAL) exhibit superior performance compared to their conventional counterparts. However, the performance of NFs-PoE was not significantly better than that of Proj-MPPI, and in all experiments, TT-PoE consistently outperformed NFs-PoE by a substantial margin. In the subsequent sections, we will delve deeper into the reasons behind this performance, explaining the advantages of using TT-distribution over NFs models and identifying its limiting factors.

\subsection{TT-distribution vs Normalizing flow}
The rationale behind choosing normalizing flows (NFs) over other generative approaches like autoencoders \citep{schmidhuber2015deep} or generative adversarial networks (GANs) \citep{goodfellow2014generative} lies in the NFs' unique ability to map samples from the latent space to the main space and vice versa. Such bidirectional mapping is crucial when combining multiple distributions in a product, ensuring they operate within the same space.

\begin{figure*}[bt]
    \centering
    \subfigure[Latent space samples]{%
        \includegraphics[height=0.29\textwidth]{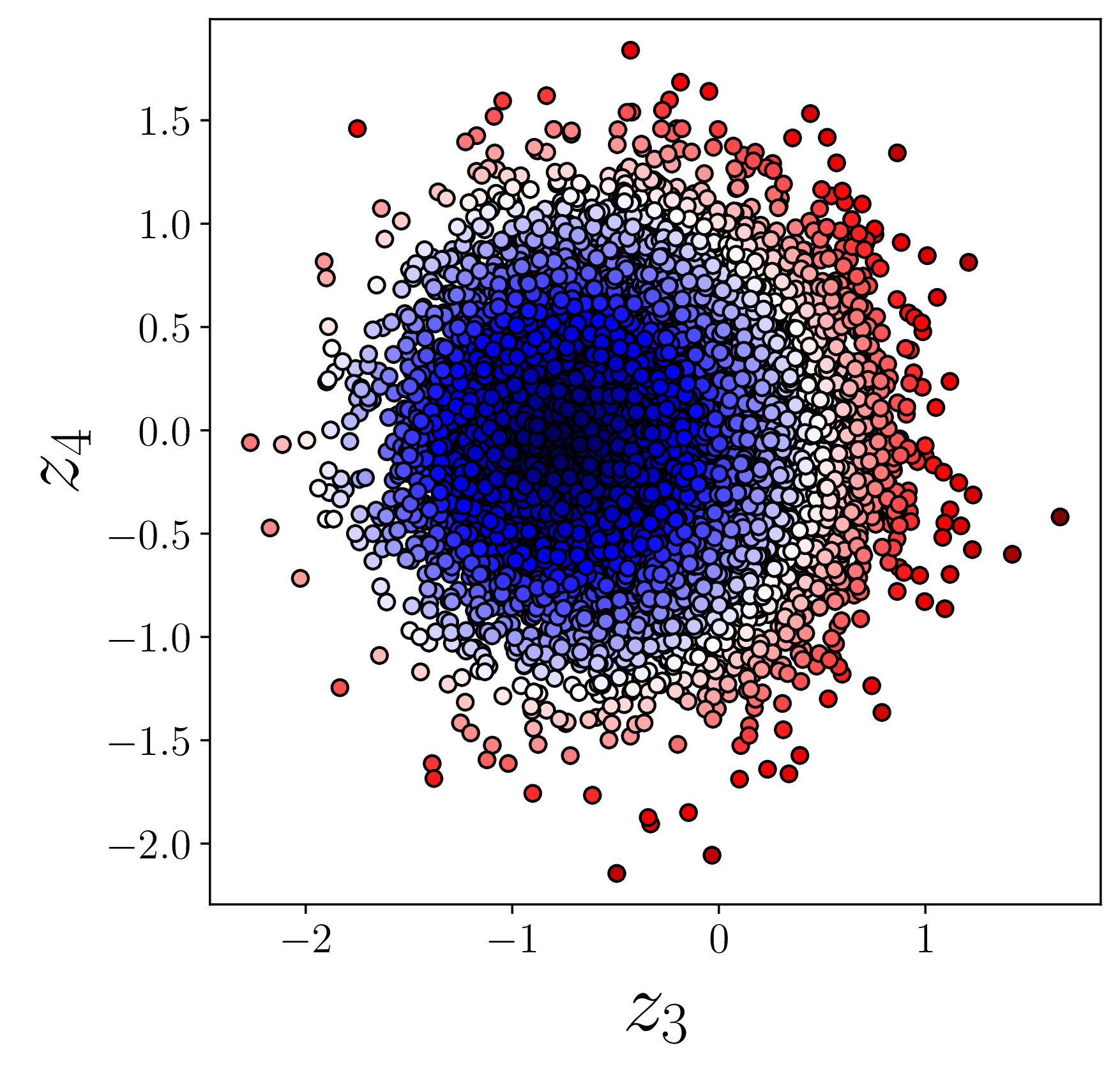}
        }
    \hfill
    \subfigure[Samples mapped to the task space]{%
        \includegraphics[height=0.29\textwidth]{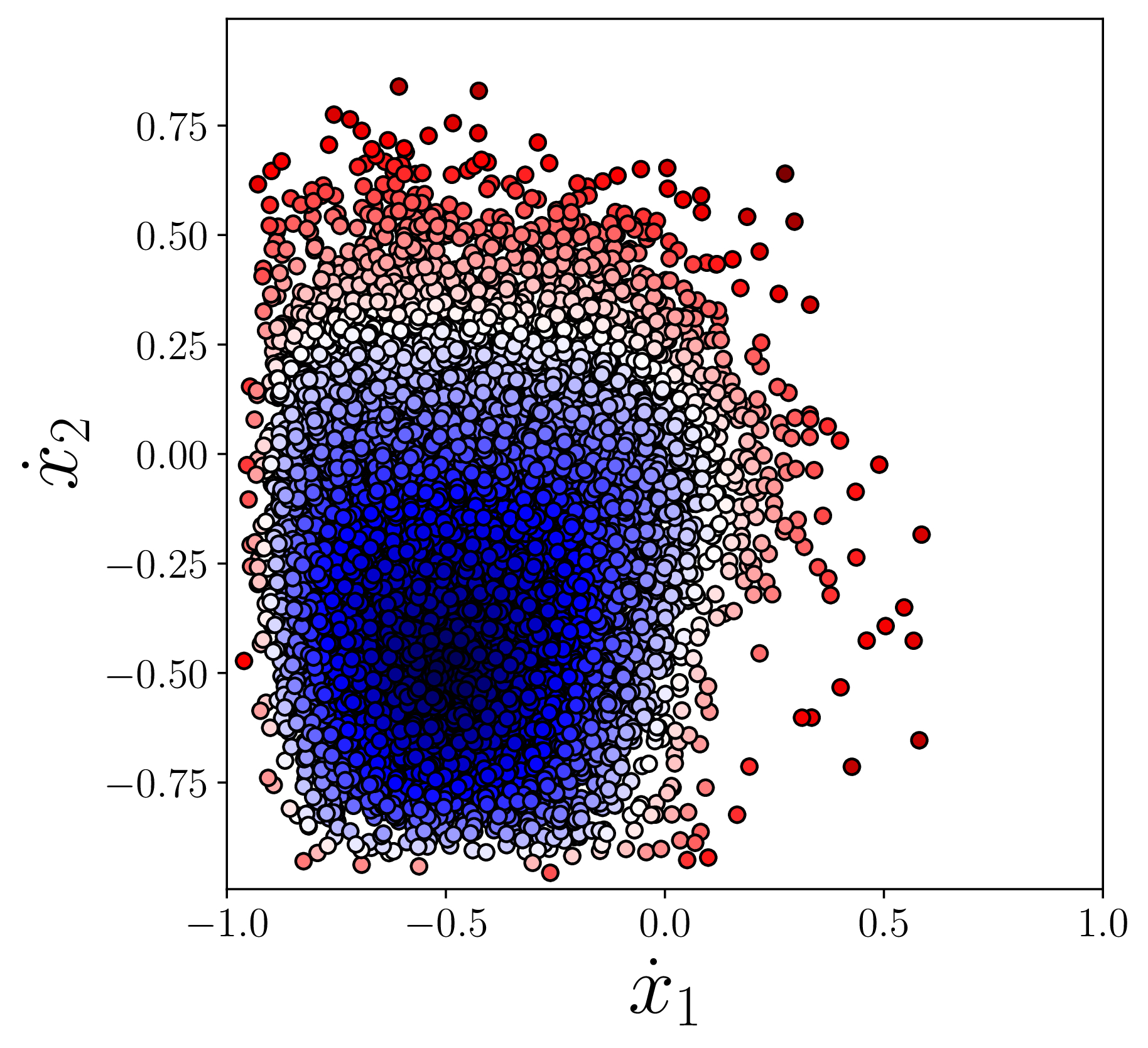}
        }
    \hfill
    \subfigure[Samples from TT-distribution]{%
        \includegraphics[height=0.29\textwidth]{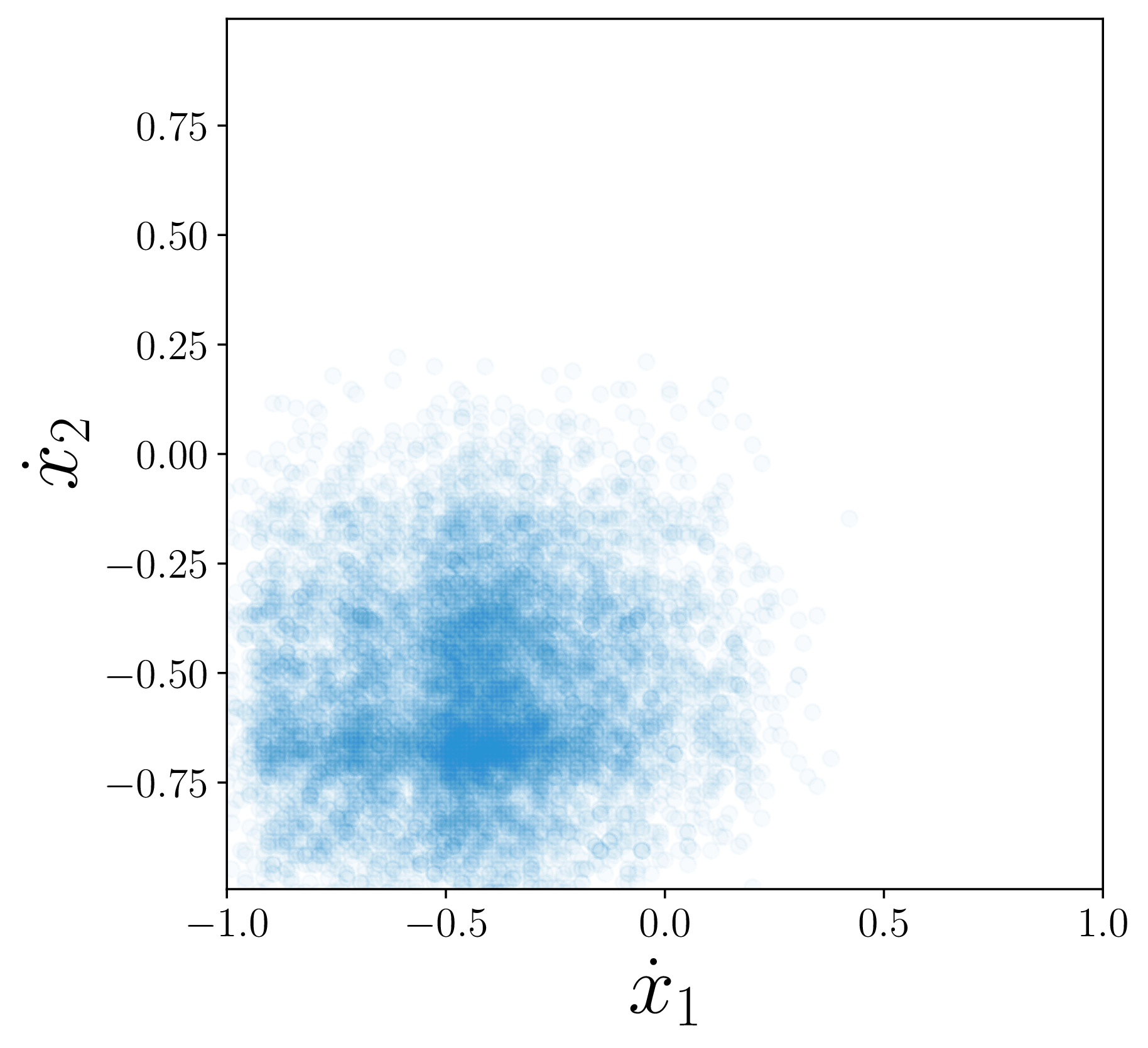}
        }
    \caption{Samples gathered with different methods. { In (a) and (b), the color gradient represents the proximity of the samples to the mean action of MPPI, with darker blue indicating closer proximity and darker red indicating greater distance. It is evident that samples generated in the latent space are distorted when translated back to the task space: they are not centered around the desired action and exhibit varying spans across dimensions despite having the same variance. Conversely, samples generated via the TT method are centered around the desired action and exhibit consistent spans in both dimensions.}}
    \label{fig:samples_comp}
\end{figure*}

Nevertheless, similar to other latent space learning methods, the acquired latent space lacks \emph{intuitiveness} \citep{losey2022learning}. In this context, \emph{intuitive} refers to the observed relation between data points in the latent space, not necessarily aligning with their distance in the main spaces. For instance, when two proximate data points exhibit similar distances to an objective point in latent space, while in task space, one is positioned near the nominal solution and the other is distant. This phenomenon is illustrated in Fig. \ref{fig:samples_comp}, obtained from a model trained for the PNGRID task, where the nominal action is presumed to be $[-0.5,-0.5]$ at state $[-1,-0.9]$, and the Gaussian model utilized for the MPPI method possesses identical variance in both directions. At this point, only velocity constraints (ranging between -1 and 1) must be active, with no other constraints in effect.

To generate samples from normalizing flows (NFs), the process involves mapping the nominal state-action back into the latent space (referred to as nominal-latent point), assuming a Gaussian distribution around it, combining it with the learned feasible distribution for the task, and sampling from the resulting distribution. The nominal latent point is depicted in dark blue in Fig. \ref{fig:samples_comp}-a, with other samples represented in distinct colors corresponding to their distance from the nominal latent point (darker blue indicates proximity, while darker red signifies greater distance). These samples are then mapped back into the task space, shown in Fig. \ref{fig:samples_comp}-b, where the colors in both images correspond, i.e. each sample maintains the same color in both representations. Notably, all samples fall within the -1 to 1 range in both directions, representing the action limits (see Appendix for experiment details).

Observing the task space samples, it is evident that they are not entirely clustered around the nominal action $[-0.5,0.5]$. Additionally, the distribution of samples in both directions appears dissimilar. This discrepancy arises due to the nonlinear function employed to revert the data back to the task space, which may have varying effects in different directions. This suggests that the covariance matrix in the latent space must be meticulously chosen to ensure the desired distribution upon reverting the data to the task space. From the perspective of obtaining random samples, this might not seem like a significant issue, as there are indeed samples close to the nominal solutions. However, the bias introduced in NFs can render the MPPI solution inefficient, which we believe is a key factor contributing to the inferior performance of this method compared to our proposed method, as well as to the MPPI method in certain cases. 
    
In contrast, samples obtained from the TT-distribution, as illustrated in Fig. \ref{fig:samples_comp}-c, are centered around the desired nominal point and exhibit a consistent distribution in both directions. This is attributed to the fact that in the TT-distribution, the product of the two distributions is executed in the task space without the need to map the data to the latent space and back. This preserves the original distribution shape, contributing to increased efficiency. An alternative methodology involves the application of techniques like GMMs directly in the task space. While this approach holds promise in addressing the previously mentioned issue, a challenge arises from its parametric nature, a topic that will be further explored in subsequent discussions.

Another challenge associated with normalizing flows is that, despite enhancing the probability of positive samples in the latent space, it does not entirely eliminate the probability of negative samples. This stems from the method relying on the Gaussian distribution, a parametric distribution wherein the probability values decrease smoothly without abruptly reaching zero. This issue is also prevalent in other distribution learning methods such as GMMs. To illustrate this point, we model the distribution of the feasible set of the PNGRID area using GMM. Fig. \ref{fig:pnr_grid_env}-b displays approximately 10,000 samples generated from the GMM model, which employed 400 Gaussian components. The results show that the model fails to efficiently capture the boundaries, compared to TT-distribution samples depicted in Fig. \ref{fig:pnr_grid_env}-d. Additionally, it is noteworthy that the training process for the GMM model required significantly more time: approximately 21 minutes, compared to around 5 minutes for normalizing flows and just milliseconds for TT-SVD.

This issue also affects the product distribution when combining it with the one obtained from MPPI. If the two distributions (optimality and feasibility) do not have overlapping high-probability areas, such that the nominal trajectory is close to regions of infeasibility (low-probability areas), the system will consider the low-probability areas of both distributions as possible candidates.

In contrast, the TT approach mitigates this problem to a greater extent. As a non-parametric distribution model, TT does not require probabilities to change smoothly, allowing it to capture sharp transitions in the distribution more effectively. Furthermore, the TT model takes both positive and negative samples into account, aiming to assign a probability of zero to the infeasible areas and one to the feasible areas. Consequently, it acts as a filter to eliminate the possibility of sampling from an infeasible distribution. This means that after combining the two distributions, the infeasible areas will ideally have zero probability, regardless of the characteristics of the optimality distribution.

\subsection{Low-Rank Representation}
As highlighted in prior works such as \citep{Shetty23}, one of the features that enhance the efficiency of TT methods is their ability to represent high-dimensional tensors in low-rank formats. The rank of a tensor can be understood analogously to that of a matrix, defined as the number of independent fibers in the tensor. This low-rank representation plays a pivotal role in our process for several reasons.

First, it allows the entire tensor to be stored in a significantly more memory-efficient manner. Second, techniques such as TT-cross \citep{savostyanov2011fast} exploit this property to learn the TT representation of a tensor without requiring access to all its elements, reducing computational overhead. Third, this low-rank structure enables faster and more efficient sampling strategies (compared to sampling directly from the whole tensor), an advantage leveraged extensively in our proposed method. Finally, it allows operations like tensor products to be executed directly at the core level, bypassing the need to reconstruct or manipulate the full tensor.

To clarify why low-rank representations make such operations efficient, it is essential to examine the relationship between tensor rank and the separability of a function's variables. When a function is converted into a tensor representation, each edge in the resulting tensor corresponds to one argument of the function. Without delving into excessive detail, it can be demonstrated that the rank of a tensor representing a function is inversely related to the separability of the function's variables. For instance, a fully separable function, such as $f(x,y) = f_x(x)f_y(y)$, results in a rank-1 tensor, irrespective of the specific forms or complexity of $f_x(\cdot)$ and $f_y(\cdot)$. This separability is also evident at the core level, where each tensor core corresponds to a distinct argument of the function. This property was crucial to our approach. When combining the optimality and feasibility distributions, we manipulated only the cores associated with the action commands, leaving other cores unchanged.

\subsection{Computation time} As outlined in Alg.~\ref{alg:process}, our method adds only two extra operations (lines 3–4) and modifies the sampling step (line 5) compared to MPPI. Table \ref{tab:computation_time} shows how these changes affect computation time. Although MPPI achieves faster overall runtimes, our TT-sampling times are on par with other baselines, making it suitable for online planning.

In its current Python implementation, TT‑sampling scales linearly with the number of action variables—e.g., the whole‑body obstacle‑avoidance task uses 3.5× more—because it relies on a Python library (i.e., tntorch) with a for‑loop inside, and lacks multithreading. By contrast, MPPI can leverage highly optimized native libraries and its state‑independent Gaussian distribution to improve the computation time significantly. By migrating performance‑critical TT sections to C/C++ and enabling multithreading, we estimate TT’s sampling time could drop below 1 ms. Until those optimizations are implemented, the timings reported here reflect the current implementation. All measurements were recorded using an NVIDIA RTX 3090 GPU and a 12‑core CPU.

\begin{table*}
    \centering
    \caption{\color{black}Average computation times (in milliseconds) for each method on the PNGRID (P) and whole-body obstacle avoidance (W) tasks. MPPI’s runtimes were virtually identical across both tasks and are therefore reported once. Note that computation times for related methods (e.g., DIAL and TT-PoE-DIAL) can be inferred directly from this table. Although our proposed method is slower than MPPI, it is on par with the other techniques and, as shown in the performance tables, yields a significant improvement over MPPI.}
    
    \resizebox{0.9\textwidth}{!}{\color{black}
    \begin{tabular}{l|c|c|c|c|c|c}
         Method & Product [ms]& Conditioning [ms]& Sampling [ms] & Mapping [ms]& Projection [ms]& Total [ms]\\
         \hline
         NFs-PoE-MPPI (P)& $0.03 \pm 0.00$& - & $0.48 \pm 0.02$ & $2.50 \pm 0.07 (\times 2)$ & - & $5.51 \pm 0.16$ \\
         MPPI & - & - & $0.48 \pm 0.02$ & - & - & $\bm{0.48 \pm 0.02}$ \\
         Proj\_MPPI (P)& - & - & $0.48 \pm 0.02$ & - & $5.02 \pm 26.46$ & $5.50 \pm 26.48$ \\
         TT-PoE-MPPI (P)&  $0.02 \pm 0.00$& $0.04 \pm 0.00$& $2.67 \pm 0.02$ & - & - & $2.73 \pm 0.02$ \\
         \hdashline
         \rule{0pt}{14pt}TT-PoE-MPPI (W)&  $0.13 \pm 0.01$& $1.23 \pm 0.10$& $5.92 \pm 0.31$ & - & - & $7.28 \pm 0.42$ \\
    \end{tabular}}
    \label{tab:computation_time}
    
\end{table*}

\subsection{Limitations} 
Despite its strengths, our approach has scalability limits: while we’ve demonstrated performance on moderately high-dimensional systems (which can be sufficient for a wide range of robotic tasks), it does not extend readily to very high dimensions (e.g., over 50). Several avenues could address this. First, a multi-resolution discretization scheme (e.g., using coarse grids when the robot is far from obstacles and finer grids when it nears them) can dramatically shrink the effective discretization size and enable higher-dimensional problems. Second, leveraging cumulative learning to inform the TT-cross algorithm may accelerate convergence and improve accuracy. Finally, integrating TT models with complementary learning techniques to discover compact latent representations could further reduce dimensionality and allow TT methods to operate more efficiently in even larger state spaces.

Alternatively, recent advances in generative modeling techniques, such as diffusion-based models \citep{ho2020denoising} and energy-based models \citep{lecun2006tutorial}, present opportunities for further exploration. These methods could be extended for use within the MPC framework discussed in this work, as our PoE-MPPI approach is independent of TT methods and can be adapted to incorporate these generative approaches. In this work, we have already demonstrated the mutual benefits of combining our pipeline with DIAL-MPC (a data-free, diffusion-inspired sampler), and future work could similarly integrate other generative approaches to further improve sampling efficiency and constraint satisfaction.

It is also worth noting that while generative methods typically rely on demonstration data to train individual experts, this requirement can be challenging when exploring the entire feasibility region, as demonstrations may not sufficiently cover all possible scenarios. By contrast, the TT approach does not depend on demonstration data and instead operates on functions. This design choice has its advantages for our purpose but may also introduce limitations for certain applications. In some applications, a hybrid strategy that combines data-driven and function-based experts could offer the best of both worlds.

There are two potential avenues for enhancing the proposed method. Firstly, the system's capability for long-horizon planning could be improved by incorporating an additional expert that considers environmental contexts. These methods can be regarded as supplementary experts and seamlessly integrated with other methodologies through the products of experts method elucidated in this paper, potentially leading to increased efficiency. Secondly, recent findings \citep{Shetty24ICLR,Xue24CORL} demonstrate the applicability of TT models for solving approximate dynamic programming and deriving policy models for robots in TT form. Their approach involves selecting the most optimal or robust action from the model, but we can modify the pipeline by sampling from the TT model and incorporating additional experts, as discussed in this paper. This modification enables a more flexible pipeline, distributing tasks among different experts and mitigating the reliance on achieving convergence to the most optimal solutions.
\section{Conclusion} 
\label{sec:conclusion}
In this article, we presented a novel approach to enhance the efficiency of sampling-based model predictive control by leveraging products of experts in the solution space and discussing the introduction of a feasibility expert. We illustrated the efficacy of this concept, which differs from traditional projection methods (the \emph{sample-then-project} strategy or tangent-space-related works) by first projecting the optimality distribution into the feasible area and then sampling from it (the \emph{project-then-sample} strategy). This approach increases the likelihood of each sample being accepted and provides a more diverse set of samples within the boundaries, rather than focusing solely on the boundaries. Additionally, it is well-suited for tasks where computing the gradient function is challenging, such as non-prehensile manipulation. It also establishes a flexible framework to integrate solutions from different methods, each with distinct objectives and decision variables. This capability allows us to reshape the sampling distribution before drawing samples, enabling more targeted and efficient exploration. We demonstrated how this advantage can be leveraged across different sampling-based frameworks, including MPPI and DIAL-MPC, resulting in improved sampling efficiency and task performance in both cases.

We demonstrated that this idea can be realized using tensor train (TT) distribution models, which allow for seamless integration of different distributions, including solutions from other sampling-based methods. We compared the advantages of TT-distributions to another distribution learning method, specifically normalizing flows. The significant benefits of TT-distributions lie in their non-parametric nature, their ability to facilitate product operations with other distributions at the task level, and their straightforward sampling approaches. We illustrated these advantages across multiple tasks. 

To address the current limitations of our method in higher dimensions, future research could investigate the potential of initially learning a low-dimensional latent space and subsequently leveraging the TT-PoE method. Another promising avenue involves incorporating additional experts with additional modalities (e.g., demonstrations or other optimality distributions . Similarly, some policies learned in the field of reinforcement learning, such as \citep{Shetty24ICLR,Xue24CORL}, can be viewed as distributions and could function as experts to enhance and modify the proposed method in this study.

\section*{Declaration of Conflicting Interests}
The authors declared no potential conflicts of interest with respect to the research, authorship, and/or publication of this article.

\section*{Funding}
The authors disclosed receipt of the following financial support for the research, authorship, and/or publication of this article:
This work was supported by the State Secretariat for Education, Research and Innovation in Switzerland for participation in the European Commission’s Horizon Europe Program through the INTELLIMAN project (\href{https://intelliman-project.eu/}{https://intelliman-project.eu/}, HORIZON-CL4-Digital-Emerging Grant 101070136) and the SESTOSENSO project (\href{https://sestosenso.eu/}{https://sestosenso.eu/}, HORIZON-CL4-Digital-Emerging Grant 101070310), as well as by the China Scholarship Council (grant No. 202106230104).
\section*{ORCID iDs}
\noindent
Amirreza Razmjoo \orcidlink{0000-0003-3826-6608} \href{https://orcid.org/0000-0003-3826-6608}{\small https://orcid.org/0000-0003-3826-6608}\\ Teng Xue \orcidlink{0009-0001-7414-3958} \href{https://orcid.org/0009-0001-7414-3958}{\small https://orcid.org/0009-0001-7414-3958},\\ Suhan Shetty \orcidlink{0000-0002-7550-9368} \href{https://orcid.org/0000-0002-7550-9368}{\small https://orcid.org/0000-0002-7550-9368}, \\ Sylvain Calinon \orcidlink{0000-0002-9036-6799} \href{https://orcid.org/0000-0002-9036-6799}{\small https://orcid.org/0000-0002-9036-6799}
\bibliographystyle{SageH}
\bibliography{references}

\appendix
\section{Appendix}

\subsection{TT-Sample}

We demonstrate the application of TT-cores in the process of generating new data samples. For the sake of clarity, we omit the notation of $Z$, i.e. $Z = 1$, as it is not necessary for sampling from the specified distribution, thereby $\text{Pr}(\bm{x}) = |\bm{\mc{P}}_{\bm{x}}|, \quad \bm{x} \in \Omega_{\bm{x}}$. Using a chain of conditional distributions, we can define
\begin{multline*}
        \text{Pr}(x_1,\ldots,x_d) = \text{Pr}_1(x_1)\text{Pr}_2(x_2|x_1) \cdots \\ \cdots \text{Pr}_d(x_d|x_1,\ldots,x_{d-1}),
\end{multline*}
where
\begin{equation*}
    \text{Pr}_k(x_k|x_1,\ldots,x_{k-1}) = \frac{s_k(x_1,\ldots,x_k)}{s_{k-1}(x_1,\ldots,x_{k-1})}    
\end{equation*}
is the conditional distribution defined using the marginals
\begin{equation*}
    s_k(x_1,\ldots,x_k) = \sum_{x_{k+1}}\cdots \sum_{x_d} \text{Pr}(x_1,\ldots,x_d),    
\end{equation*}
with the assumption that $s_0 = 1$. Utilizing the definitions outlined above, we can proceed to generate samples $\bm{x}\sim \text{Pr}(\cdot)$ by sampling from each conditional distribution sequentially. Each conditional distribution pertains solely to one variable, and in the discrete scenario, it can be interpreted as a multinomial distribution, wherein
\begin{equation*}
    x_k \sim \text{Pr}_k(x_k|x_1,\ldots,x_{k-1}), \text{ } \forall k \in \{1,\ldots,d\}.
\end{equation*}

This approach entails significant computational demands, as sampling $x_k$ necessitates accessing the conditional distribution $\text{Pr}_k$, which, in turn, mandates evaluating summations across multiple variables to ascertain the marginal $s_k$, which escalates exponentially with the dimensionality of the problem. The function separability, facilitated by the TT-model, can be leveraged to address this concern. Let $\bm{\mc{P}}$ denote the discrete counterpart of the function $\text{Pr}(\cdot)$, represented as a tensor in TT format, with discretization set $\mc{X}$. If the TT model is composed of cores $(\bm{\mc{P}}^1,\ldots,\bm{\mc{P}}^d)$, then we can express
\begin{align}
\footnotesize
\begin{split}
s_k(x_1,&\ldots,x_k) = \sum_{x_{k+1}} \cdots \sum_{x_{d}} \text{Pr}(\bm{x}), \\
&\approx  \sum_{x_{k+1}}  \cdots \sum_{x_d} \bm{\mc{P}}_{\bm{x}} , \\
&= \sum_{x_{k+1}} \cdots \sum_{x_{d}} \bm{\mc{P}}^1_{:,x_1,:} \cdots \bm{\mc{P}}^k_{:,x_k,:} \bm{\mc{P}}^{k+1}_{:,x_{k+1},:} \cdots \bm{\mc{P}}^d_{:,x_d,:},\\
&=  \bm{\mc{P}}^1_{:,x_1,:} \cdots \bm{\mc{P}}^k_{:,x_k,:}\Big(\sum_{x_{k+1}} \bm{\mc{P}}^{k+1}_{:,x_{k+1},:}\Big) \cdots \Big(\sum_{x_{d}} \bm{\mc{P}}^d_{:,x_d,:}\Big),
\end{split}
\label{eq:marginal}
\end{align}
where $\sum_{x_k}\bm{\mc{P}}^{k}_{:,x_{k},:}$ is the aggregation of all matrix slices within the TT-cores. Consequently, the TT format simplifies the expensive summation into one-dimensional summations. In cases where identical summation terms recur across multiple conditionals $\text{Pr}_k$, an algorithm known as Tensor Train Conditional Distribution (TT-CD) sampling \citep{dolgov2020approximation} can be employed to efficiently obtain samples from $\text{Pr}(\cdot)$.

\subsection{Conditional TT}
\label{tt_condition}
Let assume that $\bm{x}$ be composed of two parts as $\bm{x}=(\bm{y}_1,\bm{y}_2) \in \Omega_{\bm{x}}$, where $\bm{y}_1$ and $\bm{y}_2$ refers to the first $d_1$ and the last $d - d_1$ variables in $\bm{y}$, respectively. Our goal is to find $\text{Pr}(\bm{y}_2 | \bm{y}_1)$ of the TT distribution given in \eqref{eq:tt_distribution}. 

For a fixed value $\bm{y}_1=\bm{y}_t$, we define a new TT model $\bm{\mc P}^{\bm{y}_t}$ over $\bm{y}_2$ by slicing the original cores:
\[
\bm{\mc P}^{\bm{y}_t}_{\bm{y}_2}
\;=\;
\bm{\mc P}_{(\bm{y}_t,\bm{y}_2)}
\quad\forall\;\bm{y}_2\in\Omega_{\bm{y}_2}.
\]
Equivalently, its cores become
\begin{equation}
  \label{eq:conditioned_tt_model_par}
  (\bm{\mc P}^{\bm{y}_t})^k \;=\;
  \begin{cases}
    \bm{\mc P}^k_{:,\,y_{t_k},\,:}, & k=1,\dots,d_1,\\[6pt]
    \bm{\mc P}^k,                  & k=d_1+1,\dots,d.
  \end{cases}
\end{equation}

The conditional probability then follows as
\begin{equation}
  \label{eq:tt_distribution_cond_par}
  \Pr(\bm{y}_2\mid \bm{y}_1=\bm{y}_t)
  \;=\;
  \frac{\bigl|\bm{\mc P}_{\bm{y}_2}^{\bm{y}_t}\bigr|}{Z_1}
  \quad\forall\;\bm{y}_2\in\Omega_{\bm{y}_2}.
\end{equation}

Finally, one can draw samples of $\bm{y}_2$ given $\bm{y}_1=\bm{y}_t$.
\subsection{Proj-MPPI}
For this baseline, we aimed to project each sample into a feasible step before accepting it. Specifically, after sampling an action \(\hat{\bm{u}}^i_t\) for sample \(i\) at state \(\hat{\bm{x}}^i_t\), we project the action into \(\bar{\bm{u}}^i_t = \alpha^* \hat{\bm{u}}^i_t\), where
\begin{equation}
\begin{gathered}
    \alpha^* = \underset{\alpha \in [0,1]} {\text{argmax}} \quad \alpha, \\
    \text{s.t.}\quad \bm{f}(\hat{\bm{x}}^i_{t+1},\alpha \hat{\bm{u}}^i_t) \leq \bm{0}.
\end{gathered}
\end{equation}

This formulation performs a line search to determine the optimal scaling factor \(\alpha^*\), which identifies the best feasible action in the same direction as the sampled action while ensuring that constraints are satisfied.

\subsection{NFs-PoE-MPPI}
In this study, we compared our proposed method against normalizing flows (NFs) as a baseline for learning the static distribution. As discussed in the main text, NFs offer an advantage over other generative models by learning an invertible mapping function from a latent space to the primary space. This property enables the mapping of nominal solutions from the primary space to the latent space for sampling and subsequent remapping back to the primary space. However, it was noted in the main text that defining the nominal distribution in the latent space poses challenges.

To address this issue, we adopted an approach similar to what was outlined in \citep{power2023variational} for FlowMPPI. Specifically, for half of the samples, we performed standard sampling in the primary space (similar to the MPPI method). For the remaining samples, we initially transformed the nominal mean $\bm{\mu}^x_{t+h}$ from the task space to the latent space $\bm{\mu}^z_{t+h}$, assumed a Gaussian distribution around the nominal latent mean, i.e., $\bm{z} \sim \mathcal{N}(\bm{z}|\bm{\mu}^z_{t+h},\bm{\Sigma}^z_{t+h})$, combined it with the distribution learned using NFs, and then sampled from it. Finally, the samples were mapped back to the primary space. We predefined $\bm{\Sigma}^z_{t+h}$ to prioritize either the optimality or feasibity distribution based on its value. Given that the latent feasible space distribution has an identity covariance (with zero mean), a value of $\bm{\Sigma}^z_{t+h}$ greater than 1 in a specific dimension indicates a higher priority for the feasibility distribution, while a value less than 1 prioritizes the optimality distribution. The optimal value for $\bm{\Sigma}^z_{t+h}$ can vary depending on the task.

NFs were trained using the method described by \cite{dinh2016density}. To learn the conditional distribution using NFs, we trained the mapping function for all state and action variables simultaneously, applying an identity transformation to the state variables. We employed the same amount of data as used for TT-SVD, with the parameters detailed in Table \ref{tab:learning_data}.

\begin{table*}
\centering
\caption{Hyperparameters employed for learning the feasibility distribution using both tensor train and normalizing flow methods across various tasks. Please note that the obstacle avoidance task refers to both PNGRID and the online obstacle avoidance task.}
\resizebox{\textwidth}{!}{\begin{tabular}{cc|c|c|c|c|c|c|c} \label{tab:learning_data}
      & Param.& Obstacle Avoidance & Push. Cyl. & { Push. Box} & Push. Mus. & {Sphere man.} & Sin. man. &  Whole-body\\
     \hline
     {{\multirow{2}{*}{TT}}}&{{$\#$ State Discretization}} & {{$100$}}& {{$10$}} & {{$10$}} & {{$10$}} & { $25$} & $25$ & $50$\\
     & {{$\#$ Action Discretization}} & {{$20$}}& {{$10$}} & {{$10$}} & {{$10$}} & {$10$} & $10$ & {$50$}\\
     &{{State Refinement Scale}} & {{$1$}} & {{$20$}} & {{$10$}} & {{$10$}} & {$10$} & $10$ & {$1$}\\
     &{{Action Refinement Scale}} & {{$10$}} & {{$20$}} & {{$10$}} & {{$10$}} & {$10$} & $10$ & {$1$}\\
     \hline
     \multirow{4}{*}{NFs}&$\#$ Coupling Layers & $6$& $8$ & {$8$} & $8$ &{$8$} & $6$ &
     {-}\\
     &Size of layers & $256$ & $256$ & {$256$} & $256$ & {$256$} & {$256$} & {-}\\
     &Batch size & $4096$ & $4096$ & {$8192$} & $8192$ & {$8192$} & {$8192$} & {-}\\
     &learning rate & $0.0004$& $0.0002$& {$0.0001$} & $0.0001$ & {$0.0002$} & {$0.0004$} & {-}\\
\end{tabular}}
\end{table*}

\subsection{MPPI on the Sphere Manifold}

For the intrinsic method used in Section~\ref{sec:manifold}, we adopt the exponential map to enable sampling in the tangent space of the sphere, followed by mapping the samples back to the manifold. At each timestep, the tangent space is defined at a point $\bm{x}_t \in \mathcal{M} = \left\{ \bm{x} \in \mathbb{R}^3 \,\middle|\, \|\bm{x}\| = r \right\}$ on the sphere. Any valid action $\bm{u}_t \in \mathbb{R}^3$ must lie in this tangent space, and thus satisfy the orthogonality condition $\bm{x}_t^\top \bm{u}_t = 0$.

This tangent space is a 2D subspace of $\mathbb{R}^3$, and we generate actions $\bm{u}_t$ by first sampling in this lower-dimensional space: $\bm{u}_t^{2D} \sim \mathcal{N}(\bm{\mu}_t, \bm{\Sigma}_t)$. The mapping from the 2D tangent coordinates to the 3D ambient space is defined as
\begin{equation}
    \bm{u}_t = \bm{T}(\bm{x}_t) \bm{u}_t^{2D},
\end{equation}
where the transformation matrix $\bm{T}(\bm{x}_t) = [\bm{e}^\top_1(\bm{x}_t), \bm{e}^\top_2(\bm{x}_t)]^\top$ is constructed from an orthonormal basis of the tangent space at $\bm{x}_t$
\begin{equation}
    \bm{e}_1 = \frac{\bm{l} \times \bm{x}_t}{\| \bm{l} \times \bm{x}_t \|}, \quad
    \bm{e}_2 = \frac{\bm{e}_1 \times \bm{x}_t}{\| \bm{e}_1 \times \bm{x}_t \|}.
\end{equation}

Here, $\bm{l}$ is an arbitrary vector chosen to avoid alignment with $\bm{x}_t$, ensuring a well-defined cross product
\begin{equation}
   \bm{l} = 
   \begin{cases}
   [1, 0, 0]^\top & \text{if } |\bm{x}_t^\top [1, 0, 0]^\top| \leq 0.9, \\
   [0, 1, 0]^\top & \text{otherwise}.
   \end{cases}
\end{equation}

After sampling actions and mapping them to the task space, the next state $\bm{x}_{t+1}$ is computed using the exponential map on the sphere
\begin{equation}
\bm{x}_{t+1} = \exp_{\bm{x}_t}(\bm{v}) = 
\cos\left(\frac{\|\bm{v}\|}{r}\right) \bm{x}_t + 
\sin\left(\frac{\|\bm{v}\|}{r}\right) \frac{r\,\bm{v}}{\|\bm{v}\|},
\end{equation}
where $\bm{v} = \bm{u}_t \cdot dt$ is the scaled action vector, and $r$ is the radius of the sphere. This formulation ensures that the resulting state $\bm{x}_{t+1}$ remains on the sphere manifold.

\subsection{DIAL-mpc and TT-PoE-DIAL}
Inspired by the success of diffusion-based methods across various domains, \citet{pan2024modelbased} proposed a novel MPPI variant for trajectory optimization. In their approach, each MPPI iteration is repeated \( N_{\text{diffuse}} \) times using different noise realizations. Their results (which we also observe in our own experiments) demonstrate that this method significantly outperforms vanilla MPPI and several other baselines. Building on this idea, \citet{xue2025icra} further extended the method to a full MPC framework.

In their formulation, the noise covariance at each diffusion step is defined as
\begin{equation}\label{eq:noise_dial}
    \bm{\Sigma}^n_{t+h} = \exp\left(-\frac{N_{\text{diffuse}} - n}{\beta_1 N_{\text{diffuse}}} - \frac{H - h}{\beta_2 H}\right)\bm{I},
\end{equation}
where \( \beta_1 \) and \( \beta_2 \) are temperature parameters controlling the diffusion process and the time horizon, respectively. For simplicity, we define two decay factors: \( f_{\text{denoise}} = \exp\left(-\frac{1}{\beta_1 N_{\text{diffuse}}}\right) \) and \( f_{\text{horizon}} = \exp\left(-\frac{1}{\beta_2 H}\right) \).

Algorithm~\ref{alg:tt_dial} outlines how this diffusion-based MPPI method can be integrated into the TT-PoE pipeline. The only modifications required are the addition of lines 5 and 6, and replacing Gaussian sampling with the TT-based sampling approach described in Equation~\eqref{eq:prod_cores}.

\begin{algorithm}[t]
\caption{TT-PoE-DIAL}\label{alg:tt_dial}
\begin{algorithmic}[1]
\Statex \hspace*{-\algorithmicindent} \textbf{Require:} $\bm{x}_t, \bm{\mu}_{t, \ldots,t + H-1}, \bm{\sigma}_{t, \ldots,t+H-1}$
\Statex \hspace*{-\algorithmicindent} \textbf{Return:} $\bm{u}^*_t$
\State $\hat{\bm{x}}_{t} \gets \bm{x}_{t}$
\For{$n$ from $0$ to $N_{\text{diffuse}}$}
\For{$h$ from $0$ to $H-1$}
\State Update the covariance matrix \Comment{Eq. \eqref{eq:noise_dial}}
\State Calculate $\bm{\mathcal{P}}_{\mathcal{N}_m}$ \Comment{Eq. \eqref{eq:calc_gauss_cores}}
\State $\bm{\mc{P}}_h \gets$Product of the distributions \Comment{Eq. \eqref{eq:prod_cores}}
\State $\hat{\bm{u}}^0_{t+h},\ldots, \hat{\bm{u}}^N_{t+h} \gets $Sample $N$ actions 
\State $\hat{\bm{x}}^0_{t+h+1},\ldots, \hat{\bm{x}}^N_{t+h+1} \gets $Forward Dynamics
\EndFor
\State $c^0, \dots, c^N \gets $Calculate cost values
\State $\bm{\mu}_{t} \gets $ update the mean value \Comment{Eq. \eqref{eq:mppi_update}} 
\EndFor
\State $\bm{u}^*_t \gets \bm{\mu}_t$
\end{algorithmic}
\end{algorithm}

\subsection{Experiments}\label{sec:app_exp}
Different settings used for different methods in various tasks are listed in Table \ref{tab:exp_data}. These variables include parameters like the length of the planning horizon $H$, the covariance matrices in the task space $\bm{\Sigma}_{\text{task}}$, the covariance matrix $\bm{\Sigma}_z$ for the latent space, the temperature variable $\beta$ used in MPPI, the simulation time step $dt$, the maximum velocity $u_{\text{max}}$, and the state limits $x_{\text{max}}$. Task success is defined by meeting predefined criteria: completion within a specified number of steps $T_{\text{succ.}}$ and achieving a cost value below a predetermined threshold $c_{\text{succ.}}$. It is important to note that for the MPPI method, action commands are manually clipped to stay within feasible ranges while other methods handle this internally. 

\begin{table*}
\centering
\caption{Hyperparameters employed with different controllers. The obstacle avoidance task refers to both PNGRID and the online obstacle avoidance task.}

{\begin{tabular}{c|c|c|c|c|c|c|c} \label{tab:exp_data}
     & Obstacle Avoidance & Push. Cyl. & {Push. Box} & Push. Must. & {Sphere man.} & Sinus man. & { Whole-body}\\
     \hline
     $H$ & $15$ & $32$ & {$20$} & $20$ & {$50$} & $30$ & {$15$}\\
     $\bm{\Sigma}_{\text{task}}$ & $0.125 \bm{I}$ & $0.3 \bm{I}$ & {$0.05 \bm{I}$} & $0.05 \bm{I}$ & {$0.1 \bm{I}$} & {$0.05 \bm{I}$} & {$0.05 \bm{I}$}\\
     ${\bm{\Sigma}_z}$ & {$10 \bm{I}$} &  {$0.5 \bm{I}$} & {$0.01 \bm{I}$}& {$0.05 \bm{I}$} & {$0.2 \bm{I}$} & {$10 \bm{I}$} &{-}\\
     $\beta$ & $0.05$& $0.05$ & {$0.05$} & $0.05$ & {$0.05$} & {$0.05$} & {$0.01$}\\
     $dt [s]$ & $0.1$& $0.05$ & {$0.05$} & $0.05$ & {$0.1$} & {$0.1$} & {$0.1$}\\
     $u_{\max} [\frac{m}{s}]$ & $1$& $1$&$1$ &{$1$} &{$1$} &{$1$} & {Real Limits}\\
     $x_{\max} [m]$ & $1.25$& $1$ & {$1$}& $0.4$ & {$0.25$} & {$0.5$} & {Real Limits}\\
     $T_{{\text{succ.}}}$ & {$100$} &  {$500$} & {$500$}& {$400$} & {$400$} & {$200$} & {$25$}\\
     $c_{\text{succ.}}$  & {$10^{30}$} &  {$10^{4}$}&  {$10^{4}$} & {$10^3$}& {$3 \times 10^5$} & {$10^{30}$} & {$10^6$}\\{$N_{\text{diffuse}}$}  & { - } &   { - }&   { - } &  { - }&  { - } &  { - } & {$3$} \\{$\text{f}_{\text{denoise}}$}  & { - } &   { - }&   { - } &  { - }&  { - } &  { - } & {$0.22$} \\{$\text{f}_{\text{horizon}}$}  & { - } &   { - }&   { - } &  { - }&  { - } &  { - } & {$0.95$} \\
\end{tabular}}
\end{table*}

Upon computing the cost value for sample $i$, we proceed to normalize them using the minimum observed cost value among the samples, expressed as
\begin{equation}
    c^i = \frac{\hat{c}^i}{c_{\min}}, \quad \text{s.t.} \quad c_{\min} = \min(\hat{c}^0,\ldots,\hat{c}^N),
\end{equation}
where the notation $\hat{(\cdot)}$ denotes the values of the cost functions before normalization. Additionally, we incorporate an extra step by assuming one of the samples to have zero action. This facilitates the system to halt in case all samples become infeasible. Further explanation regarding the employed cost functions in these experiments is provided in the following.

\subsubsection{Obstacle avoidance:}
In these tasks (i.e., PNGRID and the online obstacle avoidance), our objective is to navigate towards a designated target point $\bm{p}_{\text{target}}$, while navigating around obstacles and staying within the environmental boundary.{ For addressing this task, we employed a single integrator point mass as our agent, resulting in a 4-dimensional state-action space. The TT-distribution for this function is depicted in Fig. \ref{fig:pnr_grid_env}-c for zero velocity. Notably, the region without obstacles exhibits a higher probability value compared to other areas. The occurrence of negative probabilities in some points is attributed to the imperfections in the TT-SVD approximation. While this could be mitigated by increasing the maximum rank value, which is set to 300 for this task, we observed satisfactory results with this configuration. The distribution is determined by discretizing the state space into 100 partitions and the action space into 20 partitions. However, the partition in the TT-cores corresponding to the action commands undergoes an additional refinement by interpolating between different nodes resulting in 10 times more partition. The samples generated by the model with zero velocity are illustrated in Fig. \ref{fig:pnr_grid_env}-d and can be contrasted with the samples obtained from a Gaussian mixture model (GMM) employing $K = 400$ Gaussian components, as shown in Fig. \ref{fig:pnr_grid_env}-b. The GMM model is trained with an EM algorithm using the Scikit-learn library \citep{scikit} in Python.}

To assess the efficacy of a sample, we employ a performance evaluation as
\begin{multline}\label{eq:pngrid_cost}
    \hat{c}_{\text{Obst. Avoid.}}(\hat{\bm{U}}_t) = \sum_{h=0}^{H-1} r_{t+h} \Big( 10 \; \| \hat{\bm{x}}_{t+h} - \bm{p}_{\text{target}}\|^2 + \\ 10^{30}\; \text{coll}(\hat{\bm{x}}_{t+h}) + 10^{-3} \; \| \hat{\bm{u}}_{t+h}\|^2 \Big) + \\
    10^3 \; r_{t+H} \| \hat{\bm{x}}_{t+H} - \bm{p}_{\text{target}}\|^2
\end{multline}
where $\text{coll}(\cdot)$ denotes a function that returns 1 if the point collides with an obstacle or exceeds the feasible area, and 0 otherwise. During the planning and learning phases of the feasibility distribution for the PNGRID task, we consider a 5-centimeter margin around obstacles to define collisions. In \eqref{eq:pngrid_cost}, $r_{t+h}$ serves as an indicator to determine whether the agent has previously reached the target by time step $t+h$ or not. This mechanism prevents the controller from solely aiming for the target at the final time step. We consider the target as reached if the distance between the agent and the target point falls below 5 centimeters.

{\color{black} For Expert 2, we have utilized the following feasibility distribution
\[
p_{\text{feas}}(\hat{\bm{x}}_t, \hat{\bm{u}}_t) \;=\;
\bm{1}\!\bigl\{\hat{\bm{x}}_t \in \mathcal{X},\;\hat{\bm{x}}_{t+1}\in \mathcal{X},\;(\hat{\bm{x}}_t,\hat{\bm{x}}_{t+1})\cap \mathcal{O} = \emptyset\bigr\},
\]
where, $\mathcal{X}$ denotes the allowable state‐space (inside the boundaries), $\mathcal{O}$ the obstacle set, and $\bm{1}\{\cdot\}$ is the indicator function.
}

For the online obstacle avoidance task, we utilized the same values without accounting for any collision margins. The only additional modification was that the obstacle became detectable to the solver when the robot was within 40 centimeters horizontally from the center of obstacles, which had radii of up to 40 centimeters, selected randomly.

\subsubsection{Planar-pushing:} We have modeled the robot end-effector as a 2D point mass under the control of a velocity controller. For each object, we have utilized distinct parameters, detailed in Table \ref{tab:exp_data}, and is subjected to unique cost functions, elaborated upon subsequently. {\color{black} For Expert~2, however, all objects share the same feasibility distribution
\[
p_{\text{feas}}(\hat{\bm{x}}_t,\hat{\bm{u}}_t)
=\bm{1}\Bigl\{\hat{\bm{x}}^{\text{obj.}}  \in \mathcal{X} \land \Big(\Delta_o>0
\;\lor\;
{\Delta_d}>0\Big)\Bigr\},
\]
where
\[
\begin{gathered}
\Delta_o = \|\hat{\bm{x}}^{\text{obj.}}_{t+1}-\hat{\bm{x}}^{\text{obj.}}_t\|, \\
\Delta_d = \|\hat{\bm{x}}^{\text{rob.}}_t-\hat{\bm{x}}^{\text{obj.}}_t\|-\|\hat{\bm{x}}^{\text{rob.}}_{t+1}-\hat{\bm{x}}^{\text{obj.}}_{t+1}\|.
\end{gathered}
\]
Here, \(\hat{\bm{x}}^{\mathrm{obj}}\) and \(\hat{\bm{x}}^{\mathrm{rob}}\) denote the object and robot states, respectively. This distribution assigns high probability only to samples in which the object remains within the feasible region and either (i) the object moves (\(\Delta_o>0\)) or (ii) the robot–object distance decreases (\(\Delta_d>0\)).}

\textbf{Cylinder:} The cost function designed for achieving a target point $\bm{p}_{\text{target}}$ in this task is expressed as
\begin{multline}\label{eq:cost_cylinder}
    \hat{c}_{\text{cyl.}}(\hat{\bm{U}}_t) = \sum_{h=0}^{H-1} r_{t+h+1} \Big( 10 \; \| \hat{\bm{x}}^{\text{obj}}_{t+h+1} - \bm{p}_{\text{target}}\|^2 +\\ 
    10^{-2} \; \text{coll\_o}(\hat{\bm{x}}^{\text{obj}}_{t+h+1}) + 10 \; \text{coll\_b}(\hat{\bm{x}}^{\text{obj}}_{t+h+1}) \\ 
    + \text{coll\_b}(\hat{\bm{x}}^{\text{rob}}_{t+h+1}) + 10^{-1} \;\text{SDF}^2(\hat{\bm{x}}^{\text{obj}}_{t+h+1},\hat{\bm{x}}^{\text{rob}}_{t+h+1}) \Big),
\end{multline}
where functions $\text{coll\_o}(\cdot)$ and $\text{coll\_b}(\cdot)$ determine collision with the obstacles and the boundary, respectively, returning 1 in case of collision and 0 otherwise. Additionally, $\text{SDF}(\cdot)$ calculates the signed distance value between the robot and the agent. Similarly to the obstacle avoidance experiment, $r_{t+h}$ serves as an indicator to verify whether the agent reached the target before the time step $t+h$. We define the target as reached when the distance between the agent and the target point is less than 5 centimeters.

\textbf{Pushing a Mustard Bottle and a Box:}
We use the same cost function for these two tasks which aimed at reaching a target position $\bm{p}_{\text{target}}$ and desired yaw angle $o^{\text{d}}$ as
\begin{multline}\label{eq:cost_mustard}
        \hat{c}_{\text{mus./Box}}(\hat{\bm{U}}_t) = \sum_{h=0}^{H-1} r_{t+h+1} \Big( 10 \; \| \hat{\bm{x}}^{\text{obj}}_{t+h+1} - \bm{p}_{\text{target}}\|^2  \\ + (\hat{o}_{t+h} - o^d)^2 + 10^{5} \; e^{-\frac{h}{4}} \text{coll\_b}(\hat{\bm{x}}^{\text{obj}}_{t+h+1}) \\ + 10^{5} \; e^{-\frac{h}{4}} \text{coll\_b}(\hat{\bm{x}}^{\text{rob}}_{t+h+1}) + 10 \; \|\bm{x}^{\text{rob}}_{t+h+1} - \bm{x}^{\text{obj}}_{t+h+1}\|^2 \Big) \\ + 10^3 \; r_{t+H} \Big(\| \hat{\bm{x}}^{\text{obj}}_{t+H} - \bm{p}_{\text{target}}\|^2 + (\hat{o}_{t+h} - o^d)^2\Big) \\ + s \; \text{Ang\_rel}(\hat{\bm{x}}^{\text{obj}}_t,\hat{\bm{x}}^{\text{rob}}_t,\hat{\bm{x}}^{\text{rob}}_{t+H},\theta^d),
\end{multline}
where $\hat{o}$ denotes the orientation of the object, and the remaining notation is the same as in the previous task. Additionally, for boundary cost considerations, a margin of 10 and 16 centimeters is assumed during the training and planning phases, respectively. The key distinction between \eqref{eq:cost_mustard} and \eqref{eq:cost_cylinder} lies in the final term, which evaluates the degree of rotation of the robot around the object compared to its initial position. Due to potential issues arising from the object geometry causing the robot to become stuck on one side (i.e., $s=1$), this term is introduced to encourage the robot to rotate around the object by approximately $\theta^d = \pm \frac{\pi}{2}$ when such circumstances occur. The value of $s$ is determined by
\[
     s = \left\{\begin{array}{ll}
        1, & \text{if}\; \exp\big(-|\text{dist}^{\min}_{t+H} - \text{dist}_t|\big) > 0.96\\
        0, & \text{otherwise}\\
        \end{array}\right. ,
\]
where $\text{dist}_{t+h}$ represents the distance of the object to the target (disregarding orientation) at time step $t+h$, and the superscript $\min$ indicates the minimum distance among the samples. It is important to note that this cost is applied to all samples, penalizing instances where the system remains on one side of the object.

In this experiment, the task is deemed complete if the object is within a distance of less than 10 centimeters from the target, and the orientation difference from the desired angle is less than 0.2 radians. 

\subsubsection{Tube Tracking:} In this task, our objective is to control the robot end-effector's position within a predefined area. In the sphere manifold scenario, the area is bounded by two spheres: the inner sphere has a radius of 15 cm, while the outer sphere has a radius of 20 cm. For the sinusoidal wave, the area is defined as the region inside \(\{(x,y,z) \mid z \in 0.1 \sin(5y\pi) \pm 0.03, \quad x \in [0.2, 1.0], \quad y \in [-0.5, 0.5]\}\). During both the learning and planning phases, collision detection incorporates a 1-cm margin to ensure a safety buffer.

The cost function designed for this task is expressed as
\begin{multline}\label{eq:man_surf_cost}
    \hat{c}_{\text{Manifold}}(\hat{\bm{U}}_t) = \sum_{h=0}^{H-1} r_{t+h} \Big(10^2\; \| \hat{\bm{x}}_{t+h} - \bm{p}_{\text{target}}\|^2 + \\ 10^{4} \; \text{coll}(\hat{\bm{x}}_{t+h}) + 10^{-3} \; \| \hat{\bm{u}}_{t+h}\|^2 \Big) + \\
    10^2 \; r_{t+H} \| \hat{\bm{x}}_{t+H} - \bm{p}_{\text{target}}\|^2,
\end{multline}
where the variables and terms retain the same definitions as previously described.

For Expert~2, we employ the following feasibility distribution
\[
p_{\text{feas}}(\hat{\bm{x}}_t, \hat{\bm{u}}_t)
= \bm{1}\!\Bigl\{
\hat{\bm{x}}_t \in \mathcal{X},\;
\hat{\bm{x}}_{t+1} \in \mathcal{X},\;
\hat{\bm{x}}_t,\hat{\bm{x}}_{t+1} \in \mathcal{M}_{\epsilon}
\Bigr\},
\]
where
\[
\mathcal{M}_{\epsilon}
= \{\,\bm{x} \mid \mathrm{dist}(\bm{x},\mathcal{M}) \le \epsilon\}
\]
is an \(\epsilon\)-margin around the manifold \(\mathcal{M}\). This ensures that both the current and next states lie within the admissible state space \(\mathcal{X}\) and remain close to \(\mathcal{M}\) within the specified tolerance.

\subsubsection{Whole‐body Obstacle Avoidance:}
In this experiment, we control a Franka Emika robot in the presence of randomly placed spherical obstacles.  The optimality distribution cost is 
\begin{multline}\label{eq:whole_body_cost}
\hat{c}_{\mathrm{wb}}(\hat{\bm{U}}_t)
= \sum_{h=0}^{H-1} r_{t+h}\Bigl(
10\,\|\hat{\bm{x}}_{t+h}-\bm{p}_{\mathrm{target}}\|^2\\
+10^6\,\mathrm{coll\_b}(\hat{\bm{x}}_{t+h})
+10^6\,\exp\bigl(-\tfrac{\mathrm{dist}(\hat{\bm{x}}_{t+h})}{0.01}\bigr)
\\+10\,\|\hat{\bm{u}}_{t+h}\|^2
\Bigr)
+10^3\,r_{t+H}\|\hat{\bm{x}}_{t+H}-\bm{p}_{\mathrm{target}}\|^2,
\end{multline}
where all quantities are in joint space, \(\mathrm{dist}(\cdot)\) is computed via the Robot Distance Field \citep{Li24ICRA}, and obstacles are five spheres of radius \(0.2\,\mathrm{m}\).

The feasibility expert uses a soft‐penalty distribution:
\[
p_{\text{feas}}\bigl(\bm{x}^o_{1:N_o},\,\bm{x}_t,\,\bm{u}_t\bigr)
= \prod_{\tau\in\{t,t+1\}}
\Bigl(1 - \exp\bigl(-\tfrac{\mathrm{dist}(\bm{x}_\tau)}{\lambda}\bigr)\Bigr),
\]
where \(\lambda=0.5\).  
This definition assigns low probability to states near obstacles, ensuring both the current and next robot configurations remain safely distant from all obstacles.

\subsection{Real-robot Experiment}
For the real robot experiment, we adapted the pipeline initially developed for simulating the pushing of a mustard bottle. However, we made an adjustment by reducing the number of horizons to 10. As a safety precaution, we constrained the robot's speed to one-twentieth of its simulated counterpart, with $u_{\text{max}} = 0.05$ and $\bm{u}^{\text{real}} = \frac{\bm{u}^{\text{sim}}}{20}$. Additionally, each command is executed 20 times, pushing the object to the same target point as planned. After each set of executions, the robot refreshed its action commands and iterated the process until reaching the target.

In this experiment, we employed a 7-axis Franka robot, which we controlled using a Cartesian velocity controller operating at 1000 Hz. For the visual feedback, we utilized a RealSense D435 camera, operating at 30 Hz. Detailed results and observations from these experiments are available in the accompanying videos.

\end{document}